\pdfoutput=1
\documentclass[11pt]{article}

\usepackage[preprint]{acl}

\usepackage{times}
\usepackage{latexsym}
\usepackage[T1]{fontenc}
\usepackage[utf8]{inputenc}
\usepackage{microtype}
\usepackage{inconsolata}

\usepackage{booktabs}
\usepackage{amsfonts}
\usepackage{amsmath}
\usepackage{nicefrac}
\usepackage{xcolor}
\usepackage{graphicx}
\usepackage{multirow}
\usepackage{enumitem}
\usepackage{listings}
\usepackage{subcaption}
\usepackage{placeins}

\lstdefinestyle{prompt}{
  basicstyle=\footnotesize\ttfamily,
  breaklines=true,
  breakindent=0pt,
  breakautoindent=false,
  breakatwhitespace=false,
  columns=fullflexible,
  keepspaces=true,
  frame=single,
  framerule=0.2pt,
  xleftmargin=0.5em,
  xrightmargin=0.5em,
  captionpos=t
}

\title{MindEdit-Bench: Benchmarking Object-Level Counterfactual Spatial Reasoning in VLMs from In-the-Wild Photos}

\author{%
  \textbf{Leyuan Yu$^{1,2,\ast}$, \;
    Xiao Tang$^{1,3,\ast}$, \;
    Minghao Liu$^{1,\ast}$, \;
    Xinyuan Li$^{1,2}$} \\[2pt]
  \textbf{Xiaokai Bai$^{2}$, \;
    Sheng Zhou$^{2}$, \;
    Qunshu Lin$^{1,\dagger}$, \;
    Weihao Xuan$^{4}$, \;
    Naoto Yokoya$^{4}$} \\[4pt]
  \normalfont
  $^{1}$ZODA \quad
  $^{2}$Zhejiang University \quad
  $^{3}$Tongji University \quad
  $^{4}$The University of Tokyo \\[2pt]
  {\small \textsuperscript{$\ast$}Equal contribution.\quad
    \textsuperscript{$\dagger$}Corresponding author.}
}

\begin{document}

\maketitle

\begin{abstract}
Benchmarks for vision--language models (VLMs) mostly test \emph{observational} spatial reasoning: models describe relations already visible in the input. Existing what-if tasks typically vary the observer while keeping the scene fixed. Can VLMs instead predict the consequences of hypothetically moving or rotating an object? We introduce \textbf{MindEdit-Bench}\footnote{Dataset: \url{https://huggingface.co/datasets/ZODAOfficial/MindEdit-Bench}.}, a benchmark of six spatial reasoning tasks built from three-photo smartphone triplets of newly captured indoor scenes via an automatic in-the-wild 3D scene-graph extraction pipeline. Four tasks probe perception and perspective transformation over observed structure; two new tasks, \textbf{L4 (spatial editing)} and \textbf{L5 (cross-view visibility editing)}, probe object-level counterfactual reasoning, where correct answers are absent from all input images. Each question provides 8--24 structured answer choices, enabling answer-letter-level diagnosis of spatial and fallback errors. The benchmark covers 120 private indoor scenes not drawn from public datasets, reducing public-data pretraining-overlap risk. Across 15 VLMs on 1{,}003 human-verified questions, task-wise mean VLM accuracy is only 8\%--31\%, versus 81\%--97\% human majority-vote accuracy. The pooled human--best-VLM gap is 53\,pp, with at least 39\,pp on every task. The structured answer space further reveals non-uniform failures, including weaker camera-depth-axis inference and fallback behavior on difficult visibility-editing cases.
\end{abstract}

\section{Introduction}
\label{sec:intro}

\begin{figure*}[t]
  \centering
  \includegraphics[width=\linewidth]{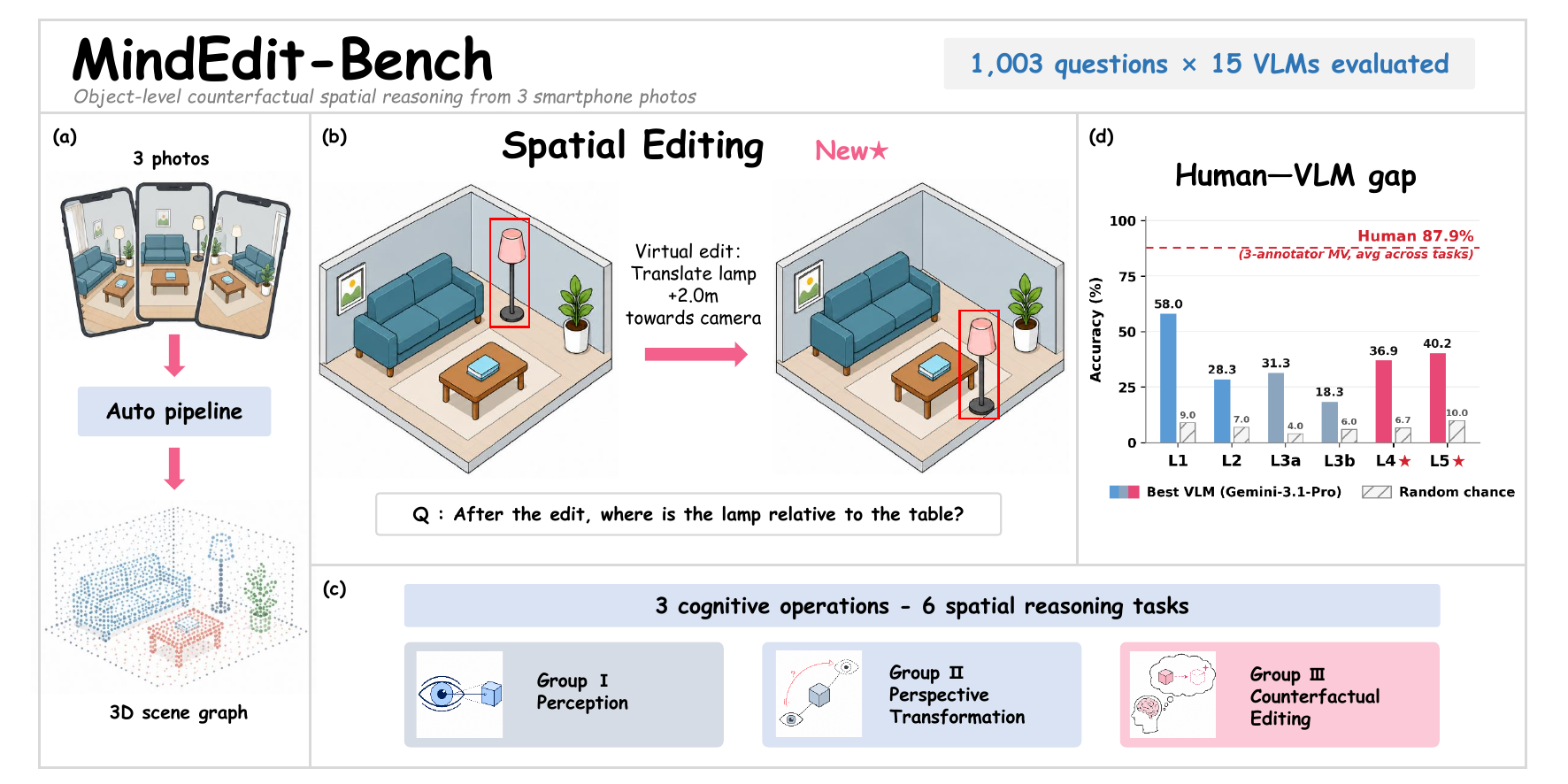}
  \caption{\textbf{MindEdit-Bench overview.}
  MindEdit-Bench probes object-level counterfactual spatial reasoning in VLMs, built from three-photo smartphone triplets sampled from privately captured indoor scenes.
  \textbf{(a)} An automatic feed-forward pipeline reconstructs an object-centric 3D scene graph.
  \textbf{(b)} An L4 spatial-editing question whose correct answer is absent from all input photos.
  \textbf{(c)} The six tasks (L1--L5) span three cognitive operations: perception, perspective transformation, and counterfactual editing.
  \textbf{(d)} Across all six tasks, best-VLM accuracy is far below the three-annotator human majority vote; the two new counterfactual editing tasks (L4 and L5) are marked with stars.}
  \label{fig:teaser}
\end{figure*}

Spatial reasoning is central to cognitive science, computer vision, and embodied AI. In vision--language models (VLMs), spatial-reasoning evaluation has progressed from single-image QA to multi-image, video, and holistic benchmark suites~\citep{spatialvlm2024, spatialrgpt2024, mmsibench2025, viewspatialbench2025, vsibench2025, spatialscore2025, easi2025, spatialtree2026}. Yet most existing benchmarks remain observational: they ask models to describe spatial relations already present in the input. This leaves a distinct question unanswered: can VLMs predict what follows when the spatial configuration itself is changed?

Object-level counterfactual spatial reasoning targets this missing case. It requires a model to represent a scene, apply a hypothetical intervention to an object, and infer the resulting spatial consequences. Existing VLM what-if benchmarks cover two narrower settings: observer-level tasks vary the viewer while keeping the scene fixed~\citep{sqa3d2023, mindcube2026}, and image-editing benchmarks modify pixels to evaluate generation fidelity rather than spatial reasoning~\citep{spatialedit2026}. The object-level case, where an object is hypothetically translated or rotated while the observer remains fixed, has not been systematically evaluated.

We introduce \textbf{MindEdit-Bench}, a six-task spatial reasoning benchmark built from three-photo smartphone triplets of privately captured indoor scenes. Four tasks test perception and perspective transformation over observed 3D structure (L1--L3b), while two test object-level counterfactual reasoning. L4 spatial editing asks how an object relation changes after virtual translation or rotation; L5 cross-view visibility editing asks which virtual edit would make an object appear or disappear in another view. Since L4/L5 answers appear in none of the input images, they remove direct 2D visual cues by construction and require reasoning over an internal 3D scene representation.

MindEdit-Bench is built on 120 newly and privately captured indoor scenes that were not part of any public dataset before our evaluation. An automatic in-the-wild pipeline generates candidate questions from three smartphone photos by extracting an object-centric 3D scene graph using feed-forward depth estimation, segmentation, and cross-view object association, without professional scanning, dense video capture, or per-scene optimisation; the final evaluation set is human-audited for correctness. Each question provides 8--24 structured answer choices, enabling direction, distance, motion-axis, operation-type, and fallback errors to be diagnosed from the final answer letter alone.

We evaluate 15 VLMs on 1{,}003 human-verified questions. Across six task types, mean VLM accuracy ranges from 8\% to 31\%, while three-annotator human majority vote reaches 81--97\%, leaving a pooled human--best-VLM gap of 53\,pp. Beyond aggregate accuracy, MindEdit-Bench shows that VLMs do not yet robustly maintain and manipulate the 3D scene representation needed for object-level counterfactual edits. Their failures are not uniform: the diagnostic answer grid reveals weaker spatial inference along the camera depth axis than along lateral/image-plane directions, while demanding cross-view visibility-editing cases elicit fallback behavior in weaker models.

Our contributions are threefold:
\begin{itemize}[leftmargin=*,nosep]
  \item \textbf{C1.} We present an end-to-end feed-forward engine that extracts object-centric 3D scene graphs from raw indoor photos and generates structured spatial-reasoning questions.
  \item \textbf{C2.} We introduce, to our knowledge, the first benchmark for object-level counterfactual spatial editing in VLMs, through L4 spatial editing and L5 cross-view visibility editing, whose answers appear in none of the input images.
  \item \textbf{C3.} We design 8--24 structured answer choices that lower random-guess baselines and enable letter-level diagnosis of direction, distance, motion-axis, operation-type, and fallback errors, without chain-of-thought analysis.
\end{itemize}

\section{Related Work}
\label{sec:related}

\paragraph{Spatial reasoning benchmarks for VLMs.}
\label{sec:rw-benchmarks}
VLM spatial-reasoning benchmarks have expanded along the input axis, from single-image QA~\citep{spatialvlm2024, spatialrgpt2024} to multi-image~\citep{mmsibench2025, viewspatialbench2025}, video~\citep{vsibench2025}, and holistic benchmark suites~\citep{spatialscore2025, easi2025, spatialtree2026}. They expose persistent gaps in perspective-taking, multi-view integration, and broad spatial understanding, and show that VLMs can exploit textual or pixel-level shortcuts instead of consistently grounding answers in 3D geometry~\citep{spatialeval2024}. However, these benchmarks largely keep the scene configuration fixed: models reason about relations or viewpoints over what is already observed. MindEdit-Bench instead evaluates object-level counterfactual reasoning: models must infer the spatial consequences of hypothetically moving or rotating an object.

\paragraph{Hypothetical spatial reasoning and cognitive foundations.}
\label{sec:rw-hypothetical}
Counterfactual spatial reasoning is closely related to cognitive accounts of mental models, viewpoint-independent spatial inference, and mental rotation~\citep{johnsonlaird1983mental,tversky1993cognitive,shepard1971mental}. More broadly, mental simulation has been argued to support mechanical and physical reasoning~\citep{hegarty2004mechanical,battaglia2013simulation}; \citet{gerstenberg2024csm} formalises counterfactual reasoning as building a generative mental model, imposing an intervention, and reading off its consequences. Existing VLM what-if benchmarks instantiate this idea at two narrower granularities: observer-level changes, where the viewer moves or reorients while the scene stays fixed~\citep{sqa3d2023,mindcube2026}, and image-editing tasks, which evaluate generation fidelity rather than spatial reasoning~\citep{spatialedit2026}. MindEdit-Bench introduces the object-level case: an object is translated or rotated while the observer remains fixed, requiring the model to re-derive spatial relations and cross-view visibility in a configuration absent from all input images.

\paragraph{In-the-wild spatial data pipelines.}
\label{sec:rw-pipelines}
Spatial benchmark construction often trades off geometric precision against collection cost. Scan-based resources rely on dedicated capture hardware~\citep{dai2017scannet}, while recent in-the-wild pipelines such as Holi-Spatial~\citep{holispatial2026} use dense video and per-scene 3D Gaussian Splatting to generate large QA sets at substantial compute cost. MindEdit-Bench occupies a lower-cost point in this design space: from only three handheld smartphone photos per scene, a feed-forward pipeline extracts an object-centric 3D scene graph and generates candidate questions without professional scanning, dense multi-view capture, or per-scene optimisation. The resulting loss in geometric precision is mitigated through sweet-zone filtering, distance margins, and human audit.

\section{MindEdit-Bench}
\label{sec:benchmark}

MindEdit-Bench evaluates VLMs on six spatial reasoning tasks constructed from three handheld smartphone photos of an indoor scene (Section~\ref{sec:engine}). The six tasks are organised by task characteristics into \textbf{three groups}. \textbf{Group~I Perception} comprises L1 cross-view object relations and L2 camera motion, which can in principle be answered within the camera reference frame of the input image and serve as the observational baseline. \textbf{Group~II Perspective Transformation} comprises L3a single-view perspective transformation and L3b cross-view perspective transformation, which require the model to reason from an imagined viewpoint different from any input camera. \textbf{Group~III Counterfactual Editing} comprises L4 spatial editing and L5 cross-view visibility editing, which ask about configurations produced by a hypothetical modification of the scene and its cross-view consequences. Table~\ref{tab:taxonomy} summarises the six types; choice counts differ across task types and each carries a different random baseline, which we report alongside every main result.

\begin{table*}[!tbp]
  \centering
  \footnotesize
  \setlength{\tabcolsep}{5pt}
  \caption{Six task types in MindEdit-Bench, organised into three groups by task characteristics. ``2D shortcut'' indicates whether pixel-based heuristics can yield correct answers without 3D reasoning.}
  \label{tab:taxonomy}
  \begin{tabular}{lllcl}
    \toprule
    Group & ID & Task & \ Choices & 2D shortcut \\
    \midrule
    \multirow{2}{*}{\textbf{I} \, Perception} & L1 & Cross-View Object Relations & 10--12 & partial \\
     & L2 & Camera Motion & 12--15 & partial (rotation) \\
    \midrule
    \multirow{2}{*}{\textbf{II} \, Perspective Transformation} & L3a & Single-View Perspective Transformation & 8--24 & none \\
     & L3b & Cross-View Perspective Transformation & 8--24 & none \\
    \midrule
    \multirow{2}{*}{\textbf{III} \, Counterfactual Editing} & L4 & Spatial Editing & 14--16 & \textbf{none} \\
     & L5 & Cross-View Visibility Editing & 8--12 & \textbf{none} \\
    \bottomrule
  \end{tabular}
\end{table*}

\subsection{Task Taxonomy}
\label{sec:taxonomy}

\paragraph{Group I Perception.}
These tasks ask about the observed scene and camera motion without requiring an imagined viewpoint or edited scene configuration, and therefore serve as observational baselines. We also note whether each task admits partial 2D shortcuts, since prior work shows that VLMs often exploit pixel-level or textual cues instead of consistently grounding answers in 3D geometry.

\textbf{L1: Cross-View Object Relations}. Given three images of the same scene from different viewpoints, the model identifies which pair of objects has the smallest 3D surface distance. Since no single image contains all objects and relevant depth relations, the task requires integrating spatial information across views. 2D shortcuts are partially effective: objects that appear close in 2D pixel space are often close in 3D, providing a weak but usable signal.

\textbf{L2: Camera Motion}. Given two images, the model predicts camera motion as a translation direction (forward / backward / left / right / negligible) plus a rotation direction (yaw left / yaw right / pitch up / pitch down / negligible). This task tests whether the model treats images as viewpoints of one 3D scene. 2D shortcuts can help with rotation, e.g., leftward scene shift implies rightward yaw, but not with depth-axis translation, which appears only as subtle 2D scaling.

\paragraph{Group II Perspective Transformation.}
These tasks require the model to reason from an imagined viewpoint different from any input camera. Because the implied reference frame no longer coincides with the image frame, direct image-plane heuristics are no longer reliable.

\textbf{L3: Perspective Transformation}. Given three reference objects A, B, and C, the model imagines standing at A facing B, then judges C's direction and distance in this new frame. This requires transforming from the camera's egocentric frame to an A-centered frame where A$\rightarrow$B is ``forward''. 2D shortcuts are ineffective by construction: an object on the image right may become ``in front of you'' after facing B, breaking the link between image-plane and spatial directions. L3 has two variants: \textbf{L3a (single-view)}, where A, B, and C appear in one image, and \textbf{L3b (cross-view)}, where they appear across images, adding multi-view integration to perspective transformation.

\begin{figure*}[t]
  \centering
  \includegraphics[width=\linewidth]{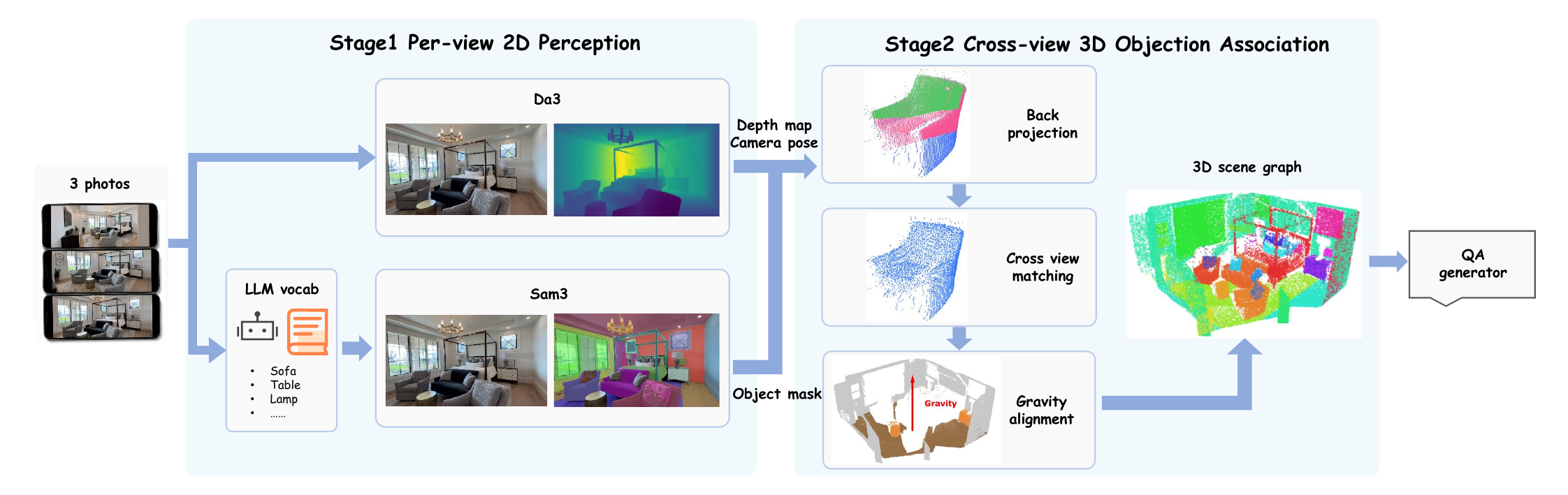}
  \caption{\textbf{3D Scene-Graph Extraction Pipeline.}
  The two-stage feed-forward pipeline takes three smartphone photos and produces an object-centric 3D scene graph that feeds the question generator (\S\ref{sec:benchmark}).
  \textbf{Stage~1: per-view 2D perception.} DA3 produces a metric depth map and camera pose for each photo; SAM3, prompted by an LLM-generated object vocabulary (e.g., sofa, table, lamp), produces per-view object masks.
  \textbf{Stage~2: cross-view 3D object association.} Per-view masks and depths are back-projected into 3D, point clusters across the three views are matched, and the resulting scene is aligned to gravity.}
  \label{fig:pipeline}
\end{figure*}

\paragraph{Group III Counterfactual Editing.}
These tasks probe object-level counterfactual reasoning: a hypothetical translation or pivot rotation is applied to a chosen object, and the model must reason about the resulting scene configuration. Because the post-edit configuration appears in none of the input images, direct visual lookup from the observed images is ruled out by construction.

\textbf{L4: Spatial Editing}. An object~A is virtually translated along one of four horizontal directions (left, right, toward camera, away from camera) by a scene-dependent distance, or rotated around a second reference object's vertical axis by $90^{\circ}$, $180^{\circ}$, or $270^{\circ}$. The model then judges A's direction and distance relative to a reference object~C. Because the answer exists in none of the input images, the model cannot directly match the query to an observed visual configuration; it must mentally execute the transformation and re-evaluate spatial relations in the modified scene.

\textbf{L5: Cross-View Visibility Editing}. The model is given two views (source and target) of the same scene, an object~A, and a list of candidate transformations of~A: translations along a direction by a fixed distance, or rotations around another reference object. The task has two variants. In the appear variant, A is visible in the source view but absent from the target view; the model must identify the transformation that would bring A into the target view. In the disappear variant, A is visible in both views; the model must identify the transformation that would remove A from the target view. Solving either variant requires mentally applying the candidate transformation, projecting A's new position into the target camera, and accounting for occlusion by other scene objects. An option ``none of the above'' may also be correct.

Detailed per-task generation algorithms, distractor sampling rules, and implementation parameters for all six tasks are in Appendix~\ref{app:question-gen}.

\subsection{Structured Answer Design}
\label{sec:quality}

Each question provides 8--24 multiple-choice options, reducing random-guess baselines to roughly $4$--$13\%$ rather than $25\%$ in standard 4-choice formats. The larger answer space lets distractors be structured along meaningful spatial axes. In L2, each option pairs a translation direction with a rotation direction, so the answer letter encodes both motion components. In L3/L4, options form an eight-direction $\times$ distance-bin grid, so the letter encodes both predicted direction and distance. In L5, options cover translation edits, rotation edits, and a ``none of the above'' fallback, so the letter encodes the selected operation type or fallback response. This enables letter-level diagnosis of direction, distance, motion-axis, operation-type, and fallback errors without chain-of-thought parsing.


\section{Data Engine}
\label{sec:engine}

MindEdit-Bench is constructed by a feed-forward pipeline that takes three casually captured smartphone photos as input and produces structured spatial reasoning questions. We deliberately trade scene-graph completeness and geometric precision for accessibility. Our pipeline does not aim to recover every object in the scene; instance segmentation is followed by a quality filter (size, confidence) that retains only clearly-segmentable scene objects suitable for anchoring questions. The three views are taken with sufficient view overlap to support cross-view 3D association (Stage~2 of Section~\ref{sec:pipeline}) while preserving enough viewpoint diversity for the perspective-transformation and cross-view visibility tasks. We privately captured 120 indoor scenes ($\sim$810 triplets) using consumer smartphones; these images were not part of any public dataset before our evaluation, reducing the risk of overlap with VLM pretraining data. Capture protocols are in Appendix~\ref{app:pipeline}; the object filter is in Appendix~\ref{app:question-gen}. End-to-end pipeline quality is quantified by human audit in Section~\ref{sec:pipeline-validity}.

\subsection{3D Scene-Graph Extraction Pipeline}
\label{sec:pipeline}

We organise the pipeline in two stages, illustrated in Figure~\ref{fig:pipeline}: per-view 2D perception (Stage~1) and cross-view 3D object association (Stage~2). Concrete hyperparameters for each stage are deferred to Appendix~\ref{app:pipeline}.

\paragraph{Stage 1: Per-view 2D perception.}
Each input photo is processed independently to produce three per-view primitives. An LLM (Gemini~3.1~Pro) examines all three photos in a single API call to produce a unified object vocabulary, so that the same physical object is not renamed across views. Feed-forward monocular depth and camera intrinsics/extrinsics are recovered by DA3~\citep{depthanything3}, whose feed-forward design matches our three-photo input. Classical Structure-from-Motion pipelines, by contrast, rely on sufficient multi-view feature correspondences and can be brittle under our sparse three-photo capture setting. Open-vocabulary instance segmentation via SAM3~\citep{sam3} is guided by the Stage-1 vocabulary; we apply a within-class non-maximum suppression to merge over-segmented parts (e.g., a table split into top and legs), with cross-class merges forbidden.

\begin{figure*}[t]
  \centering
  \includegraphics[width=\linewidth]{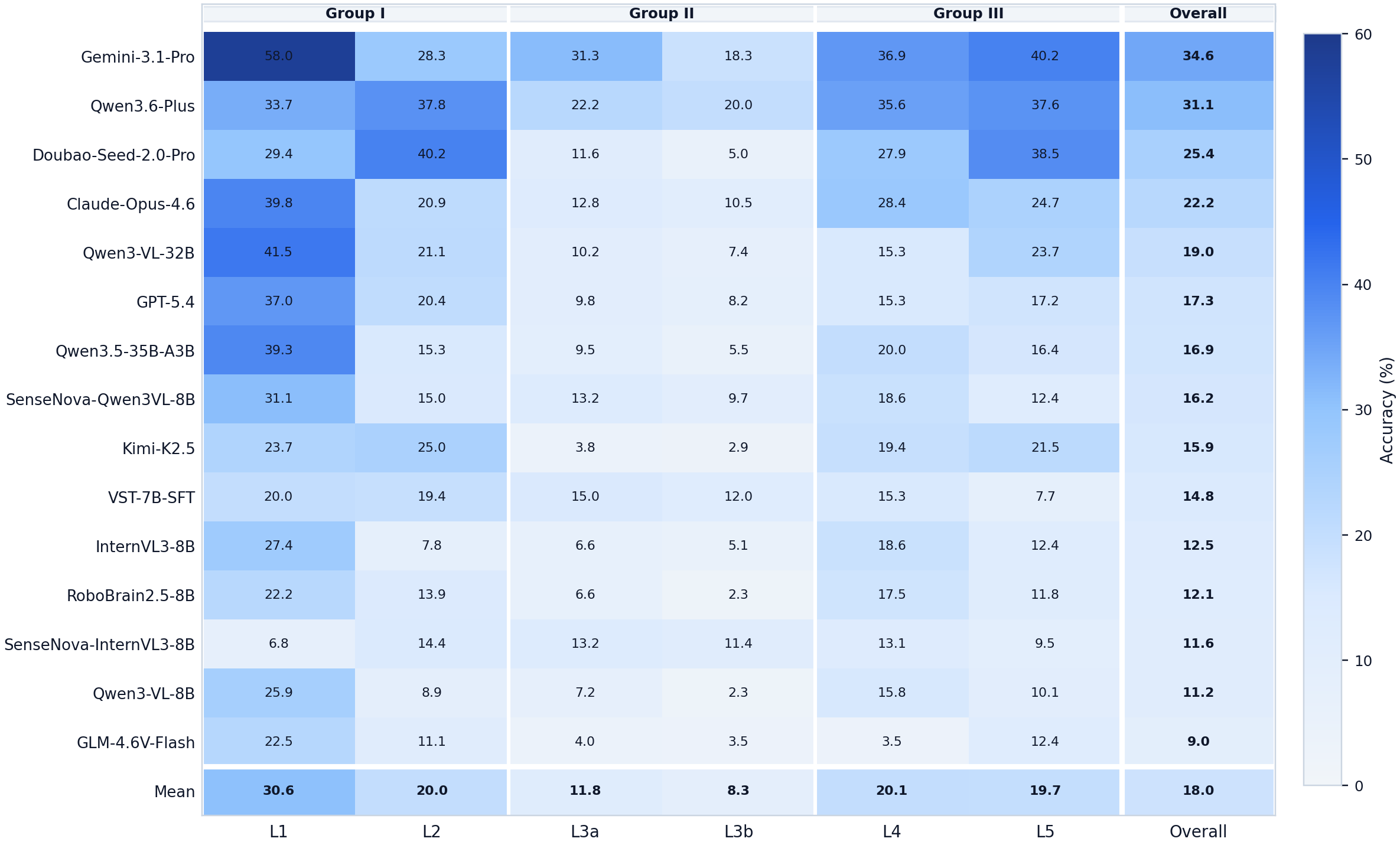}
  \caption{Per-model accuracy scoreboard across six task types. Rows are 15 VLMs ranked by overall accuracy; cells show accuracy (\%) and the same quantity is encoded by colour. The rightmost column shows each model’s pooled overall score over all examples, while the bottom row shows the per-task mean across models. L1 is the only task with a clear signal for top models, whereas L3b remains close to chance across the board.}
  \label{fig:heatmap}
\end{figure*}

\paragraph{Stage 2: Cross-view 3D object association.}
The per-view perception of Stage~1 produces mask sets that are independent across views. Stage~2 recovers this cross-view correspondence in 3D. For each view in turn, every segmented mask is back-projected into the world frame, yielding a 3D point cluster. If the cluster matches an existing 3D object of the same semantic class, its points are merged into that object; otherwise the cluster seeds a new object. Association is scored by a composite of three geometric signals: volumetric overlap, containment, and centroid proximity. Planar structures (floor, wall, ceiling) are handled by a separate coplanarity rule. Finally, the world frame is gravity-aligned by detecting horizontal planar surfaces (floor, ceiling, table-top), fitting a normal vector to each via PCA, and voting on these normals to recover the vertical direction. The resulting Z-axis alignment is a prerequisite for horizontal direction labels in downstream direction- and editing-related tasks. The filtered object pool is then consumed by the question generator (Section~\ref{sec:benchmark}), together with pairwise object distances and Z-buffer-based cross-view visibility primitives: distance bins for L1, L3, and L4, and visibility flips for L5.

\subsection{End-to-End Pipeline Validity}
\label{sec:pipeline-validity}

Of 1{,}172 candidate questions produced by the pipeline, 758 (64.7\%) are \emph{direct pass}: both the question stem and the geometry-derived answer key after language refinement were independently verified as correct. 245 (20.9\%) are \emph{answer-fixed}: the stem was unambiguous but the answer key was corrected by annotators (typically a wrong direction or distance bin). The remaining 169 (14.4\%) are \emph{rejected} due to referent-ambiguous stems. The 1{,}003-question benchmark reported in Section~\ref{sec:experiments} combines the direct-pass and answer-fixed subsets; the per-task breakdown is reported in Appendix~\ref{app:audit}. This audit indicates that the combination of structured answer design (Section~\ref{sec:quality}) and question-generation filters, including direction sweet zones and distance-bin margins (Appendix~\ref{app:question-gen}), supports benchmark construction under the residual noise of our feed-forward pipeline.

\section{Experiments}
\label{sec:experiments}

\subsection{Setup}
\label{sec:setup}

\paragraph{Main evaluation.}
To establish where current VLMs stand on object-level counterfactual spatial reasoning, we evaluate 15 VLMs on all 1{,}003 questions ($\approx 170$ per task) in a zero-shot setting, with three runs each. The API tier (6 models) comprises GPT-5.4~\citep{openai2026gpt54}, Gemini-3.1-Pro-Preview~\citep{deepmind2026gemini31pro}, Claude-Opus-4-6~\citep{anthropic2026opus46}, Doubao-Seed-2.0-Pro~\citep{bytedance2026seed2}, Qwen3.6-Plus~\citep{qwen2026qwen36plus}, and Kimi-K2.5~\citep{moonshot2026kimik25}; the open-source tier (9 models) comprises Qwen3-VL series at 8B / 32B~\citep{qwen2025qwen3vl}, InternVL3-8B~\citep{zhu2025internvl3}, SenseNova-SI-1.1~\citep{sensenova2026si} (two variants on Qwen and InternVL backbones), RoboBrain2.5~\citep{baai2026robobrain25}, VST-7B-SFT~\citep{yang2025vst}, GLM-4.6V-Flash~\citep{zai2025glm46v}, and Qwen3.5-35B-A3B~\citep{qwen2026qwen35}. We score only the final answer letter. Because option counts vary (8--24) across task types, the per-task random-guess baseline (4--13\%) is shown alongside every result for interpretation. Full evaluation settings and prompts are provided in Appendix~\ref{app:eval-protocol}.

\begin{figure}[t]
  \centering
  \includegraphics[width=\columnwidth]{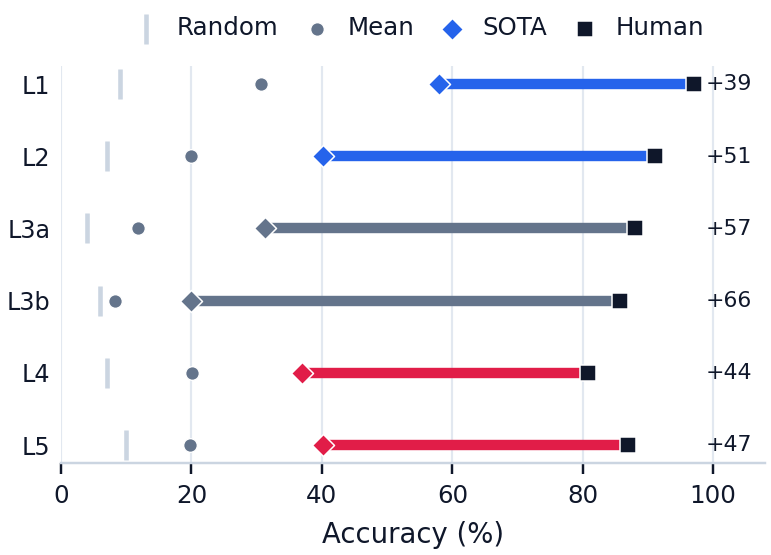}
  \caption{Task-level human--SOTA gaps. Coloured segments show the gap from the best model to human majority vote; random chance and the 15-model mean are shown as context.}
  \label{fig:task-gap}
\end{figure}

\paragraph{Human baseline.}
To estimate human performance under the same input-and-choice format, three human annotators independently answered all 1{,}003 questions, seeing the same three images and answer choices as the models. Krippendorff's $\alpha = 0.82$ (nominal, overall) indicates substantial inter-annotator agreement; we use majority vote as the human label.

\paragraph{Answer permutation.}
To verify that model answers are not driven by letter position, we shuffle the option-letter ordering and re-evaluate Gemini-3.1-Pro-Preview, Qwen3-VL-32B, and SenseNova-SI-1.1-Qwen3-VL-8B three times each.

\paragraph{Chain-of-thought prompting.}
We further probe four representative models (Gemini-3.1-Pro-Preview, GPT-5.4, Qwen3-VL-32B, Doubao-Seed-2.0-Pro) with standard chain-of-thought (CoT) prompting, implemented as a generic step-by-step reasoning instruction, to assess how much this prompting strategy narrows the human--VLM gap; full results are in Appendix~\ref{app:cot}.

\subsection{Results}
\label{sec:results}

\paragraph{Main results.}
\label{sec:main}

All six tasks remain substantially below human accuracy: the strongest VLM, Gemini-3.1-Pro-Preview, achieves only 34.6\% pooled overall accuracy over all 1,003 questions, far below the 87.9\% human majority vote. As a check against possible Gemini-related prompt-style bias, the best non-Gemini overall model, Qwen3.6-Plus, reaches only 31.1\% pooled overall accuracy, still 56.8\,pp below the human majority vote. Figure~\ref{fig:heatmap} shows per-model per-task accuracy across all 15 VLMs (full matrix of run means in Table~\ref{tab:main-full}, Appendix~\ref{app:full-results}); Figure~\ref{fig:task-gap} summarises the task-level random baseline, VLM mean, best VLM, and human majority-vote accuracy on a shared scale. We report per-task gaps to the random baseline rather than cross-task ordering, as the benchmark is not designed to support a difficulty ranking across task types with non-uniform option structure.

\paragraph{Human baseline.}
\label{sec:human-baseline}
As shown in Figure~\ref{fig:task-gap} and Table~\ref{tab:human-baseline}, per-task majority-vote accuracy ranges from $80.8\%$ to $97.0\%$, confirming that all six tasks are solvable by humans. The human-vs-SOTA gap, however, is at least 39\,pp on every task, with a pooled overall gap of 53.3\,pp, exposing a substantial VLM-vs-human capability gap that cannot be attributed to task infeasibility.

\begin{table}[t]
  \centering
  \footnotesize
  \setlength{\abovecaptionskip}{2pt}
  \setlength{\belowcaptionskip}{2pt}
  \setlength{\tabcolsep}{4pt}
  \caption{Human vs. best VLM by task. The All row is pooled over all 1,003 examples, not the unweighted macro-average of task rows. Acc. \%, gaps pp.}
  \label{tab:human-baseline}
  \begin{tabular*}{\columnwidth}{@{\extracolsep{\fill}}lrrr@{}}
    \toprule
    Task & Human & SOTA & Gap \\
    \midrule
    L1 & 97.0 & 58.0 & +39.0 \\
    L2 & 91.1 & 40.2 & +50.9 \\
    L3a & 88.0 & 31.3 & +56.7 \\
    L3b & 85.7 & 20.0 & +65.7 \\
    L4 & 80.8 & 36.9 & +43.9 \\
    L5 & 87.0 & 40.2 & +46.8 \\
    \midrule
    \textbf{All} & \textbf{87.9} & \textbf{34.6} & \textbf{+53.3} \\
    \bottomrule
  \end{tabular*}
\end{table}

\paragraph{Answer permutation.}
\label{sec:permutation}
Table~\ref{tab:permutation} shows that, at the overall-accuracy level, the difference between the main run and three option-letter shuffles stays within 1.4\,pp on all three probed models; at the task level, the largest fluctuation reaches only 6.9\,pp (Qwen3-VL-32B on L2; full per-task breakdown and visualization in Table~\ref{tab:permutation-per-task} and Figure~\ref{fig:permutation}, Appendix~\ref{app:permutation-per-task}). This indicates that, for the probed models, overall accuracy is not primarily driven by answer-letter position.

\begin{table}[t]
  \centering
  \footnotesize
  \setlength{\abovecaptionskip}{2pt}
  \setlength{\belowcaptionskip}{2pt}
  \setlength{\tabcolsep}{2.5pt}
  \caption{Answer permutation. Acc. \%, deltas pp.}
  \label{tab:permutation}
  \begin{tabular*}{\columnwidth}{@{\extracolsep{\fill}}lrrrl@{}}
    \toprule
    Model & Main & Perm. & $\Delta_{\mathrm{all}}$ & max $\Delta_{\mathrm{task}}$ \\
    \midrule
    Gemini-3.1-pro & 34.6 & 34.7 & 0.1 & 5.0 (L3b)\\
    Qwen3-VL-32B & 19.0 & 20.4 & 1.4 & 6.9 (L2)\\
    SenseNova-Q3VL & 16.2 & 16.1 & 0.1 & 6.4 (L1)\\
    \bottomrule
  \end{tabular*}
\end{table}

\subsection{Analysis}
\label{sec:analysis}

\paragraph{Depth-axis weakness.}
Beyond aggregate accuracy, the structured answer grid allows us to attribute model errors to specific spatial dimensions without chain-of-thought analysis. On L4 spatial editing, predictions binned by the eight compass directions show a clear depth-axis weakness: pure-depth answers (\emph{far} at $0^{\circ}$ and \emph{near} at $180^{\circ}$) reach only $7.4\%$ accuracy, compared with $18.0\%$ for pure-lateral answers (\emph{left} and \emph{right}) and $24.3\%$ for mixed diagonals. This gives pure-depth answers a $-10.6$\,pp gap relative to pure-lateral answers. The same pattern appears across tasks: L3a single-view perspective transformation shows a $-5.5$\,pp depth-vs-lateral gap, and L2 camera motion shows a $-9.6$\,pp gap when translation is decomposed into forward/backward depth versus left/right lateral movement ($29.4\%$ vs.\ $39.0\%$).

When the direction is wrong, errors tend to be local rather than reversed. On L4, wrong-direction predictions cluster in the adjacent $45^{\circ}$ compass bin at $1.7\times$ the random rate, whereas $180^{\circ}$ reversals occur at only $0.6\times$ the random rate. This suggests that models often localize the answer to an approximate region but rarely flip it outright. The asymmetry is unlikely to be an artifact of the direction-bin design: each L3 and L4 question describes the eight directions as $45^{\circ}$-wide sectors, while ground-truth selection applies the same $\pm 10^{\circ}$ sweet zone around every direction; angles outside all sweet zones are excluded during question generation.

Taken together, these observations indicate that current VLMs maintain a coarse, compass-style region sense but struggle to make reliable predictions along the camera depth axis. This pattern is consistent with the hypothesis that 2D-dominant visual pretraining makes image-plane position more directly accessible than camera-depth geometry.

\paragraph{Operation difficulty and fallback behavior.}
On counterfactual editing tasks, model accuracy decreases as the required edit becomes more complex. On L4, translation operations, which require only fixed-axis displacement, reach $21.8\%$ accuracy, while pivot rotations, which require angular re-derivation, drop to $13.5\%$, a $-8.3$\,pp gap. The same translation-vs-rotation pattern appears on L5: $16.5\%$ for translation versus $8.2\%$ for rotation, again a $-8.3$\,pp gap. Within each task, translation and rotation cases share the same answer format and evaluation protocol, suggesting that the gap reflects the additional spatial transformation required by rotation rather than option-count differences.

L5 further exposes a fallback failure mode. Since the task combines object editing, cross-view projection, and occlusion reasoning, weaker models often over-select ``none of the above'': SenseNova-SI-1.1-Qwen3-VL-8B selects it $95.1\%$ of the time, InternVL3-8B $88.3\%$, and Qwen3-VL-8B $73.0\%$, far above the $14.2\%$ ground-truth frequency. This over-selection is not due to answer-letter position, as it persists under answer permutation (Section~\ref{sec:permutation}); full rates are in Appendix~\ref{app:l5-fallback}.

Together, these patterns indicate difficulty-sensitive behavior: easier edit cases yield higher accuracy from the same models, while structurally demanding visibility-editing cases elicit fallback responses from models that fail to resolve the required edit-and-project reasoning.

\section{Conclusion}
\label{sec:conclusion}

MindEdit-Bench evaluates object-level counterfactual spatial reasoning in VLMs from casually captured smartphone photos, introducing L4 spatial editing and L5 cross-view visibility editing whose answers are absent from all input images. Built with a feed-forward object-centric 3D scene-graph extraction pipeline and structured answer choices, it enables low-cost benchmark construction and letter-level diagnosis of spatial and fallback errors. Across 15 VLMs and 1{,}003 human-verified questions, current models remain far below human performance and show structured failures along the camera depth axis and on demanding cross-view visibility edits. The benchmark dataset is available at \url{https://huggingface.co/datasets/ZODAOfficial/MindEdit-Bench} under the CC BY 4.0 license.

\section*{Limitations}
\label{sec:limitations}

Because L4 and L5 use task-specific filters to exclude ambiguous angles, near-boundary distances, and operations with marginal visibility change, their absolute accuracies reflect both task demands and selection criteria. Cross-task accuracy differences should therefore be interpreted cautiously, while within-task comparisons remain informative about relative model behavior. Our feed-forward scene-graph extraction pipeline (DA3, SAM3, incremental fusion) is less precise than per-scene methods such as 3D Gaussian Splatting, trading precision for low cost; residual error is mitigated by the same filters and by human audit. For some weaker models, single-run accuracy is noisy; we report the mean over three runs, but fine-grained rankings between low-stability models should be interpreted with caution. Finally, Gemini-3.1-Pro is used for object-vocabulary construction and description refinement in our data pipeline while Gemini-3.1-Pro-Preview is also evaluated; although answer keys derive from geometry and human audit, this may introduce prompt-style familiarity, so we additionally report the best non-Gemini overall result (Section~\ref{sec:main}); the human--VLM gap remains large after excluding Gemini.

\bibliography{references}

\begin{thebibliography}{35}
\providecommand{\natexlab}[1]{#1}

\bibitem[{{Anthropic}(2026)}]{anthropic2026opus46}
{Anthropic}. 2026.
\newblock \href {https://www.anthropic.com/system-cards/} {System card: {Claude
  Opus 4.6}}.
\newblock Technical report, Anthropic.

\bibitem[{Battaglia et~al.(2013)Battaglia, Hamrick, and
  Tenenbaum}]{battaglia2013simulation}
Peter~W. Battaglia, Jessica~B. Hamrick, and Joshua~B. Tenenbaum. 2013.
\newblock Simulation as an engine of physical scene understanding.
\newblock \emph{Proceedings of the National Academy of Sciences (PNAS)},
  110(45):18327--18332.

\bibitem[{{ByteDance Seed}(2026)}]{bytedance2026seed2}
{ByteDance Seed}. 2026.
\newblock \href {https://seed.bytedance.com/en/seed2} {{Seed2.0} model card}.

\bibitem[{Cai et~al.(2025{\natexlab{a}})Cai, Wang, Gu, Pu, Xu, Wang, Yin, Yang,
  Wei, Sun, Zhou, Li, Pang, Qian, Wei, Lin, Shi, Deng, Han, Chen, Fan, Deng,
  Lu, Pan, Li, Liu, Wang, Lin, and Yang}]{sensenova2026si}
Zhongang Cai, Ruisi Wang, Chenyang Gu, Fanyi Pu, Junxiang Xu, Yubo Wang, Wanqi
  Yin, Zhitao Yang, Chen Wei, Qingping Sun, Tongxi Zhou, Jiaqi Li, Hui~En Pang,
  Oscar Qian, Yukun Wei, Zhiqian Lin, Xuanke Shi, Kewang Deng, Xiaoyang Han,
  and 10 others. 2025{\natexlab{a}}.
\newblock \href {https://doi.org/10.48550/arXiv.2511.13719} {Scaling spatial
  intelligence with multimodal foundation models}.
\newblock \emph{arXiv preprint arXiv:2511.13719}.

\bibitem[{Cai et~al.(2025{\natexlab{b}})Cai, Wang, Sun, Wang, Gu, Yin, Lin,
  Yang, Wei, Qian, Pang, Shi, Deng, Han, Chen, Li, Fan, Deng, Lu, Li, Liu,
  Wang, Lin, and Yang}]{easi2025}
Zhongang Cai, Yubo Wang, Qingping Sun, Ruisi Wang, Chenyang Gu, Wanqi Yin,
  Zhiqian Lin, Zhitao Yang, Chen Wei, Oscar Qian, Hui~En Pang, Xuanke Shi,
  Kewang Deng, Xiaoyang Han, Zukai Chen, Jiaqi Li, Xiangyu Fan, Hanming Deng,
  Lewei Lu, and 5 others. 2025{\natexlab{b}}.
\newblock \href {https://doi.org/10.48550/arXiv.2508.13142} {Holistic
  evaluation of multimodal {LLM}s on spatial intelligence}.
\newblock \emph{arXiv preprint arXiv:2508.13142}.

\bibitem[{Carion et~al.(2025)Carion, Gustafson, Hu, Debnath, Hu, Suris, Ryali,
  Alwala, Khedr, Huang, Lei, Ma, Guo, Kalla, Marks, Greer, Wang, Sun,
  R{\"a}dle, Afouras, Mavroudi, Xu, Wu, Zhou, Momeni, Hazra, Ding, Vaze,
  Porcher, Li, Li, Kamath, Cheng, Doll{\'a}r, Ravi, Saenko, Zhang, and
  Feichtenhofer}]{sam3}
Nicolas Carion, Laura Gustafson, Yuan-Ting Hu, Shoubhik Debnath, Ronghang Hu,
  Didac Suris, Chaitanya Ryali, Kalyan~Vasudev Alwala, Haitham Khedr, Andrew
  Huang, Jie Lei, Tengyu Ma, Baishan Guo, Arpit Kalla, Markus Marks, Joseph
  Greer, Meng Wang, Peize Sun, Roman R{\"a}dle, and 19 others. 2025.
\newblock \href {https://doi.org/10.48550/arXiv.2511.16719} {{SAM} 3: Segment
  anything with concepts}.
\newblock \emph{arXiv preprint arXiv:2511.16719}.

\bibitem[{Chen et~al.(2024)Chen, Xu, Kirmani, Ichter, Sadigh, Guibas, and
  Xia}]{spatialvlm2024}
Boyuan Chen, Zhuo Xu, Sean Kirmani, Brian Ichter, Dorsa Sadigh, Leonidas
  Guibas, and Fei Xia. 2024.
\newblock {SpatialVLM}: Endowing vision-language models with spatial reasoning
  capabilities.
\newblock In \emph{Proceedings of the IEEE/CVF Conference on Computer Vision
  and Pattern Recognition (CVPR)}.

\bibitem[{Cheng et~al.(2024)Cheng, Yin, Fu, Guo, Yang, Kautz, Wang, and
  Liu}]{spatialrgpt2024}
An-Chieh Cheng, Hongxu Yin, Yang Fu, Qiushan Guo, Ruihan Yang, Jan Kautz,
  Xiaolong Wang, and Sifei Liu. 2024.
\newblock {SpatialRGPT}: Grounded spatial reasoning in vision language models.
\newblock In \emph{Advances in Neural Information Processing Systems
  (NeurIPS)}.

\bibitem[{Dai et~al.(2017)Dai, Chang, Savva, Halber, Funkhouser, and
  Nie{\ss}ner}]{dai2017scannet}
Angela Dai, Angel~X. Chang, Manolis Savva, Maciej Halber, Thomas Funkhouser,
  and Matthias Nie{\ss}ner. 2017.
\newblock {ScanNet}: Richly-annotated 3d reconstructions of indoor scenes.
\newblock In \emph{Proceedings of the IEEE/CVF Conference on Computer Vision
  and Pattern Recognition (CVPR)}.

\bibitem[{Gao et~al.(2026)Gao, Li, Liu, Ji, Gong, Liao, Liu, Zhang, Yang, Xu,
  Yang, Huang, Zhang, Liu, Sun, Zhang, and Zhong}]{holispatial2026}
Yuanyuan Gao, Hao Li, Yifei Liu, Xinhao Ji, Yuning Gong, Yuanjun Liao, Fangfu
  Liu, Manyuan Zhang, Yuchen Yang, Dan Xu, Xue Yang, Huaxi Huang, Hongjie
  Zhang, Ziwei Liu, Xiao Sun, Dingwen Zhang, and Zhihang Zhong. 2026.
\newblock \href {https://doi.org/10.48550/arXiv.2603.07660} {{Holi-Spatial}:
  Evolving video streams into holistic 3d spatial intelligence}.
\newblock \emph{arXiv preprint arXiv:2603.07660}.

\bibitem[{Gerstenberg(2024)}]{gerstenberg2024csm}
Tobias Gerstenberg. 2024.
\newblock \href {https://doi.org/10.1016/j.tics.2024.04.012} {Counterfactual
  simulation in causal cognition}.
\newblock \emph{Trends in Cognitive Sciences}, 28(10):924--936.

\bibitem[{{Google DeepMind}(2026)}]{deepmind2026gemini31pro}
{Google DeepMind}. 2026.
\newblock \href {https://deepmind.google/models/model-cards/gemini-3-1-pro}
  {{Gemini 3.1 Pro} model card}.
\newblock Technical report, Google DeepMind.

\bibitem[{Hegarty(2004)}]{hegarty2004mechanical}
Mary Hegarty. 2004.
\newblock Mechanical reasoning by mental simulation.
\newblock \emph{Trends in Cognitive Sciences}, 8(6):280--285.

\bibitem[{Johnson-Laird(1983)}]{johnsonlaird1983mental}
Philip~N. Johnson-Laird. 1983.
\newblock \emph{Mental Models: Towards a Cognitive Science of Language,
  Inference, and Consciousness}.
\newblock Harvard University Press.

\bibitem[{{Kimi Team}(2026)}]{moonshot2026kimik25}
{Kimi Team}. 2026.
\newblock \href {https://doi.org/10.48550/arXiv.2602.02276} {{Kimi K2.5}:
  Visual agentic intelligence}.
\newblock \emph{arXiv preprint arXiv:2602.02276}.

\bibitem[{Li et~al.(2025)Li, Li, Wang, Yan, Zhang, Chen, Hou, Jiang, Zhang,
  Shen, Lu, and Zhuang}]{viewspatialbench2025}
Dingming Li, Hongxing Li, Zixuan Wang, Yuchen Yan, Hang Zhang, Siqi Chen,
  Guiyang Hou, Shengpei Jiang, Wenqi Zhang, Yongliang Shen, Weiming Lu, and
  Yueting Zhuang. 2025.
\newblock \href {https://doi.org/10.48550/arXiv.2505.21500}
  {{ViewSpatial-Bench}: Evaluating multi-perspective spatial localization in
  vision-language models}.
\newblock \emph{arXiv preprint arXiv:2505.21500}.

\bibitem[{Lin et~al.(2025)Lin, Chen, Liew, Chen, Li, Shi, Feng, and
  Kang}]{depthanything3}
Haotong Lin, Sili Chen, Junhao Liew, Donny~Y. Chen, Zhenyu Li, Guang Shi,
  Jiashi Feng, and Bingyi Kang. 2025.
\newblock \href {https://doi.org/10.48550/arXiv.2511.10647} {Depth anything 3:
  Recovering the visual space from any views}.
\newblock \emph{arXiv preprint arXiv:2511.10647}.

\bibitem[{Ma et~al.(2023)Ma, Yong, Zheng, Li, Liang, Zhu, and
  Huang}]{sqa3d2023}
Xiaojian Ma, Silong Yong, Zilong Zheng, Qing Li, Yitao Liang, Song-Chun Zhu,
  and Siyuan Huang. 2023.
\newblock {SQA3D}: Situated question answering in 3d scenes.
\newblock In \emph{International Conference on Learning Representations
  (ICLR)}.

\bibitem[{{OpenAI}(2026)}]{openai2026gpt54}
{OpenAI}. 2026.
\newblock \href {https://openai.com/index/gpt-5-4-thinking-system-card/}
  {{GPT-5.4} thinking system card}.

\bibitem[{{Qwen Team}(2026{\natexlab{a}})}]{qwen2026qwen35}
{Qwen Team}. 2026{\natexlab{a}}.
\newblock \href {https://huggingface.co/Qwen/Qwen3.5-35B-A3B}
  {{Qwen3.5-35B-A3B} model card}.

\bibitem[{{Qwen Team}(2026{\natexlab{b}})}]{qwen2026qwen36plus}
{Qwen Team}. 2026{\natexlab{b}}.
\newblock \href {https://qwen.ai/blog?id=qwen3.6} {{Qwen3.6-Plus}: Towards real
  world agents}.

\bibitem[{{Qwen Team, Alibaba}(2025)}]{qwen2025qwen3vl}
{Qwen Team, Alibaba}. 2025.
\newblock \href {https://doi.org/10.48550/arXiv.2511.21631} {{Qwen3-VL}
  technical report}.
\newblock \emph{arXiv preprint arXiv:2511.21631}.

\bibitem[{Shepard and Metzler(1971)}]{shepard1971mental}
Roger~N. Shepard and Jacqueline Metzler. 1971.
\newblock Mental rotation of three-dimensional objects.
\newblock \emph{Science}, 171(3972):701--703.

\bibitem[{Tan et~al.(2026)Tan, Zhou, Li, Xu, Ji, Chen, Chi, Wang, Jia, Ao, Cao,
  Chen, Li, Liu, Wang, Rong, Lyu, Zhao, Co, Li, Han, Xie, Yao, Wang, Zhang,
  Yang, Jiao, Shi, Xie, Nie, Men, Lin, Wang, Huang, and
  Zhang}]{baai2026robobrain25}
Huajie Tan, Enshen Zhou, Zhiyu Li, Yijie Xu, Yuheng Ji, Xiansheng Chen, Cheng
  Chi, Pengwei Wang, Huizhu Jia, Yulong Ao, Mingyu Cao, Sixiang Chen, Zhe Li,
  Mengzhen Liu, Zixiao Wang, Shanyu Rong, Yaoxu Lyu, Zhongxia Zhao, Peterson
  Co, and 16 others. 2026.
\newblock \href {https://doi.org/10.48550/arXiv.2601.14352} {{RoboBrain 2.5}:
  Depth in sight, time in mind}.
\newblock \emph{arXiv preprint arXiv:2601.14352}.

\bibitem[{Tversky(1993)}]{tversky1993cognitive}
Barbara Tversky. 1993.
\newblock Cognitive maps, cognitive collages, and spatial mental models.
\newblock In \emph{European Conference on Spatial Information Theory (COSIT)},
  pages 14--24.

\bibitem[{Wang et~al.(2024)Wang, Ming, Shi, Vineet, Wang, Li, and
  Joshi}]{spatialeval2024}
Jiayu Wang, Yifei Ming, Zhenmei Shi, Vibhav Vineet, Xin Wang, Yixuan Li, and
  Neel Joshi. 2024.
\newblock Is a picture worth a thousand words? delving into spatial reasoning
  for vision language models.
\newblock In \emph{Advances in Neural Information Processing Systems
  (NeurIPS)}.

\bibitem[{Wang et~al.(2025)Wang, Yin, Zhang, Zhang, Wang, Wang, Zhang,
  Chandrasegaran, Liu, Krishna, Xie, Wu, Fei-Fei, and Li}]{mindcube2026}
Qineng Wang, Baiqiao Yin, Pingyue Zhang, Jianshu Zhang, Kangrui Wang, Zihan
  Wang, Jieyu Zhang, Keshigeyan Chandrasegaran, Han Liu, Ranjay Krishna,
  Saining Xie, Jiajun Wu, Li~Fei-Fei, and Manling Li. 2025.
\newblock \href {https://doi.org/10.48550/arXiv.2506.21458} {{MindCube}:
  Spatial mental modeling from limited views}.
\newblock \emph{arXiv preprint arXiv:2506.21458}.

\bibitem[{Wu et~al.(2025)Wu, Huang, Chen, Zhang, Wang, and
  Xie}]{spatialscore2025}
Haoning Wu, Xiao Huang, Yaohui Chen, Ya~Zhang, Yanfeng Wang, and Weidi Xie.
  2025.
\newblock \href {https://doi.org/10.48550/arXiv.2505.17012} {{SpatialScore}:
  Towards comprehensive evaluation for spatial intelligence}.
\newblock \emph{arXiv preprint arXiv:2505.17012}.

\bibitem[{Xiao et~al.(2026)Xiao, Zhang, Song, Chen, Li, Jiang, Ren, Lin, Huang,
  Huang, Li, Duan, and Qi}]{spatialedit2026}
Yicheng Xiao, Wenhu Zhang, Lin Song, Yukang Chen, Wenbo Li, Nan Jiang, Tianhe
  Ren, Haokun Lin, Wei Huang, Haoyang Huang, Xiu Li, Nan Duan, and Xiaojuan Qi.
  2026.
\newblock \href {https://doi.org/10.48550/arXiv.2604.04911} {{SpatialEdit}:
  Benchmarking fine-grained image spatial editing}.
\newblock \emph{arXiv preprint arXiv:2604.04911}.

\bibitem[{Xiao et~al.(2025)Xiao, Li, Yan, Liu, Peng, Wei, Zhou, and
  Kang}]{spatialtree2026}
Yuxi Xiao, Longfei Li, Shen Yan, Xinhang Liu, Sida Peng, Yunchao Wei, Xiaowei
  Zhou, and Bingyi Kang. 2025.
\newblock \href {https://doi.org/10.48550/arXiv.2512.20617} {{SpatialTree}: How
  spatial abilities branch out in {MLLM}s}.
\newblock \emph{arXiv preprint arXiv:2512.20617}.

\bibitem[{Yang et~al.(2025{\natexlab{a}})Yang, Yang, Gupta, Han, Fei-Fei, and
  Xie}]{vsibench2025}
Jihan Yang, Shusheng Yang, Anjali~W. Gupta, Rilyn Han, Li~Fei-Fei, and Saining
  Xie. 2025{\natexlab{a}}.
\newblock \href
  {https://openaccess.thecvf.com/content/CVPR2025/html/Yang_Thinking_in_Space_How_Multimodal_Large_Language_Models_See_Remember_CVPR_2025_paper.html}
  {Thinking in space: How multimodal large language models see, remember, and
  recall spaces}.
\newblock In \emph{Proceedings of the IEEE/CVF Conference on Computer Vision
  and Pattern Recognition (CVPR)}.

\bibitem[{Yang et~al.(2025{\natexlab{b}})Yang, Zhu, Li, Huang, Yan, Zhou, Liu,
  Li, Li, Wang, Lin, and Zhao}]{yang2025vst}
Rui Yang, Ziyu Zhu, Yanwei Li, Jingjia Huang, Shen Yan, Siyuan Zhou, Zhe Liu,
  Xiangtai Li, Shuangye Li, Wenqian Wang, Yi~Lin, and Hengshuang Zhao.
  2025{\natexlab{b}}.
\newblock \href {https://doi.org/10.48550/arXiv.2511.05491} {Visual spatial
  tuning}.
\newblock \emph{arXiv preprint arXiv:2511.05491}.

\bibitem[{Yang et~al.(2025{\natexlab{c}})Yang, Xu, Xie, Yang, Li, Lin, Zhu,
  Chen, Duan, Yue, Lin, Wang, and Pang}]{mmsibench2025}
Sihan Yang, Runsen Xu, Yiman Xie, Sizhe Yang, Mo~Li, Jingli Lin, Chenming Zhu,
  Xiaochen Chen, Haodong Duan, Xiangyu Yue, Dahua Lin, Tai Wang, and Jiangmiao
  Pang. 2025{\natexlab{c}}.
\newblock \href {https://doi.org/10.48550/arXiv.2505.23764} {{MMSI-Bench}: A
  benchmark for multi-image spatial intelligence}.
\newblock \emph{arXiv preprint arXiv:2505.23764}.

\bibitem[{{Z.ai}(2025)}]{zai2025glm46v}
{Z.ai}. 2025.
\newblock \href {https://huggingface.co/zai-org/GLM-4.6V-Flash}
  {{GLM-4.6V-Flash} model card}.

\bibitem[{Zhu et~al.(2025)Zhu, Wang, Chen, Liu, Ye, Gu, Tian, Duan, Su, Shao,
  Gao, Cui, Wang, Cao, Liu, Wei, Zhang, Wang, Xu, Li, Wang, Deng, Li, He,
  Jiang, Luo, Wang, He, Shi, Zhang, Shao, He, Xiong, Qu, Sun, Jiao, Lv, Wu,
  Zhang, Deng, Ge, Chen, Wang, Dou, Lu, Zhu, Lu, Lin, Qiao, Dai, and
  Wang}]{zhu2025internvl3}
Jinguo Zhu, Weiyun Wang, Zhe Chen, Zhaoyang Liu, Shenglong Ye, Lixin Gu, Hao
  Tian, Yuchen Duan, Weijie Su, Jie Shao, Zhangwei Gao, Erfei Cui, Xuehui Wang,
  Yue Cao, Yangzhou Liu, Xingguang Wei, Hongjie Zhang, Haomin Wang, Weiye Xu,
  and 32 others. 2025.
\newblock \href {https://doi.org/10.48550/arXiv.2504.10479} {{InternVL3}:
  Exploring advanced training and test-time recipes for open-source multimodal
  models}.
\newblock \emph{arXiv preprint arXiv:2504.10479}.

\end{thebibliography}

\FloatBarrier
\appendix

%
\section{Task Examples}
\label{app:examples}

\begin{figure*}[!tbp]
  \centering
  \includegraphics[width=\linewidth]{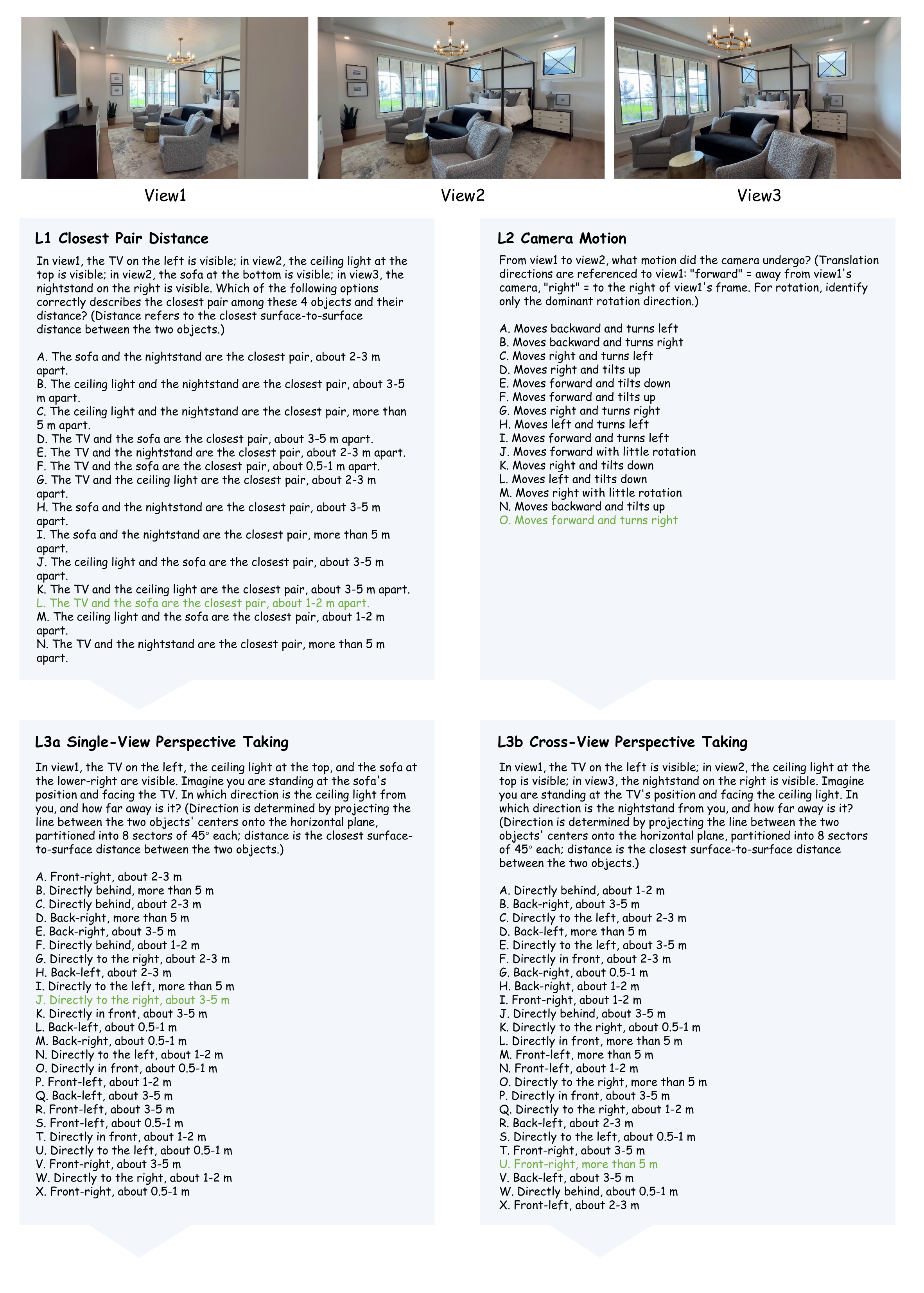}
  \caption{Representative examples for the four perception and perspective-transformation tasks (L1--L3b). Each panel shows the input photos, the question stem, the full option list, and the ground-truth answer.}
  \label{fig:qa-examples-1}
\end{figure*}

\begin{figure*}[!tbp]
  \centering
  \includegraphics[width=\linewidth]{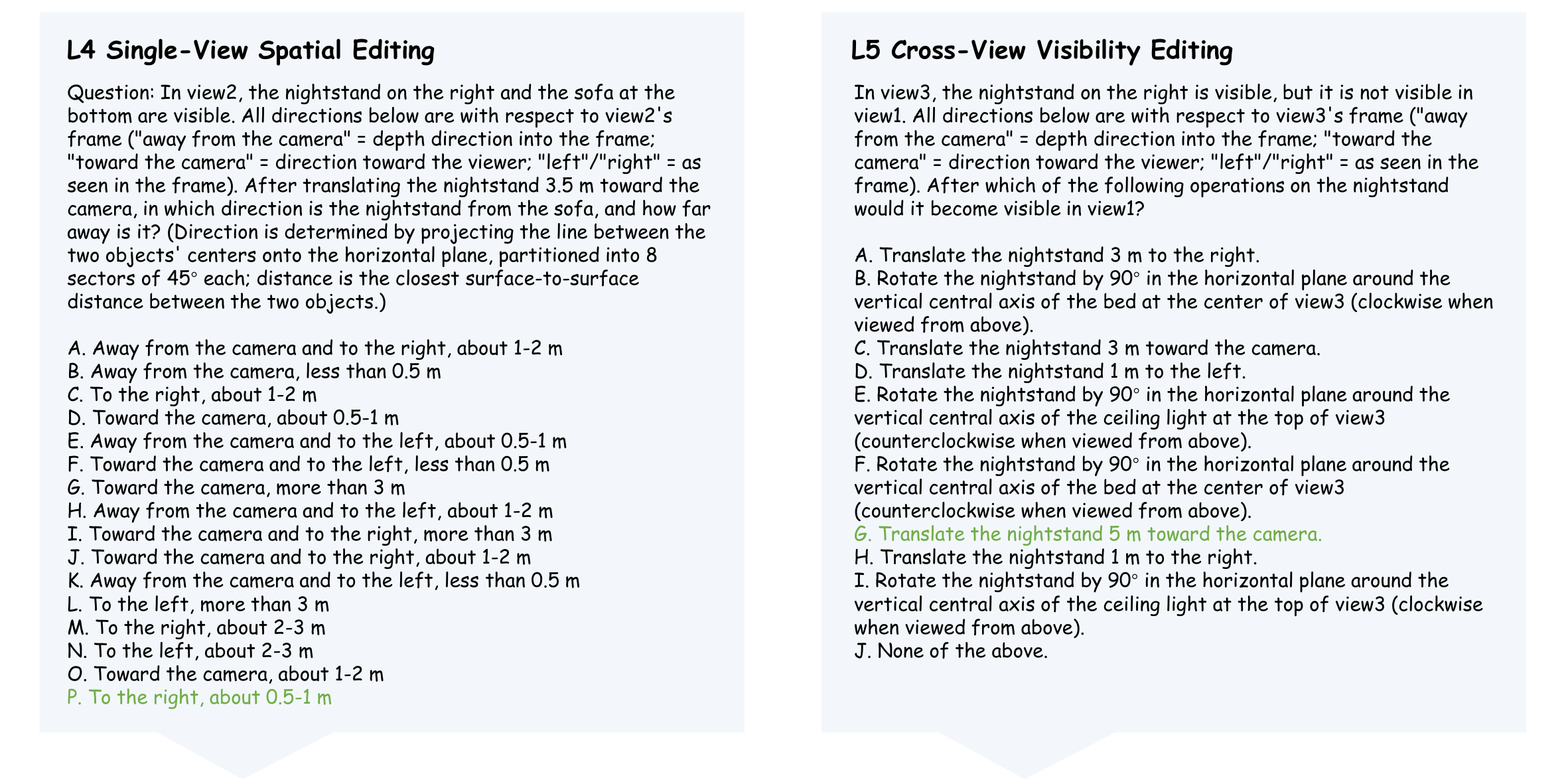}
  \caption{Representative examples for the two counterfactual editing tasks (L4 and L5). Same panel layout as Figure~\ref{fig:qa-examples-1}.}
  \label{fig:qa-examples-2}
\end{figure*}

Figures~\ref{fig:qa-examples-1} and \ref{fig:qa-examples-2} show one representative example per task type (L1--L5), including the input photos, the question stem, the full option list, and the ground-truth answer.

\section{3D Scene-Graph Extraction Pipeline}
\label{app:pipeline}

Hyperparameters of the input-photo protocol (Section~\ref{sec:engine}) and the two scene-reconstruction stages (Section~\ref{sec:pipeline}) are detailed below.

\paragraph{Input photos.}
Each scene is captured as three handheld smartphone photos taken with sufficient pairwise view overlap to support cross-view 3D association (Section~\ref{sec:pipeline}, Stage~2) while preserving enough viewpoint diversity for the perspective-transformation and cross-view visibility tasks. All scenes were privately captured by the authors and were not part of any public dataset before evaluation.

\paragraph{Stage 1 \textperiodcentered\ Per-view 2D perception.}
DA3~\citep{depthanything3} recovers per-view metric depth together with camera intrinsics and extrinsics in a single forward pass per photo. SAM3~\citep{sam3} is prompted with the Stage-1 LLM vocabulary; within-class non-maximum suppression merges masks with IoU $> 0.5$ or containment $> 0.7$, while across-class masks are never merged regardless of overlap.

\paragraph{Stage 2 \textperiodcentered\ Cross-view 3D object association.}
Per-view clusters are associated to existing 3D objects by the composite score $s = 0.4 \cdot \text{IoU}_{3\text{D}} + 0.4 \cdot \text{Containment} + 0.2 \cdot \text{CenterProximity}$ with acceptance $s \geq 0.35$ and hard floor $\text{IoU}_{3\text{D}} \geq 0.05$; weights were chosen on a held-out scene set. Planar structures use a coplanarity rule (normal-angle tolerance $< 10^{\circ}$, plane-distance tolerance $< 0.15\,\text{m}$) in place of volumetric overlap. Gravity is estimated by PCA multi-plane voting over floor, ceiling, and table-top surfaces, with ceiling-above-floor disambiguation and wall-normal consistency checks; world coordinates are then rotated so the Z-axis aligns with gravity.

\paragraph{Rendered artefacts.}
Point-cloud PLYs with 3D bounding boxes and camera frusta, together with 2D bounding-box overlays on the input images, are produced for human sanity-checking during dataset curation (Section~\ref{sec:pipeline-validity}). These are not consumed by the automated question generator.

\section{Question Generation}
\label{app:question-gen}

The question generator consumes geometric primitives derived from the reconstructed scene graph (Appendix~\ref{app:pipeline}). We detail the geometric primitives, the per-scene object pool, the per-task generation algorithms, shared parameters, and the LLM-based refinement step below.

\paragraph{Geometric primitives.}
Per-object point clouds are voxel-downsampled at 2\,cm. Pairwise surface distance: for each object pair $(A, B)$, a cKDTree is built on $B$'s points; every point of $A$ is queried against $B$ and the resulting nearest-neighbour distances are sorted ascending; we take the mean of the 5th--10th percentile slice (skipping the closest 5\% to avoid noise from overlapping points and the tail to avoid far-field noise). Cross-view visibility: Z-buffer rendering of all other objects against the projected points of the edited object gives a visibility ratio, and an edit is labelled \emph{visible} when $\text{vis\_ratio} > 0.50$, \emph{invisible} when $< 0.15$, and discarded otherwise.

\paragraph{Object pool construction.}
Before any question is generated, we apply the object filter from Stage~1 (bbox edge $0.05$--$0.80$\,m, SAM3 confidence $\geq 0.5$, plus a category exclude list to skip large fixed furniture and certain non-rigid items) to obtain a per-scene object pool. The pipeline does not aim to recover every object; the filter trades coverage for question-grade reliability.

\paragraph{Per-task generation algorithms.}

\textbf{L1: Cross-View Object Relations.} From the object pool, enumerate all object pairs $(A, B)$ that appear in at least one shared view, compute pairwise surface distance via the cKDTree-based percentile estimator. The ground-truth pair has the smallest distance. Distractor pairs are sampled with stratified distance bins: $\lceil k/2 \rceil$ pairs from the second-smallest bin, $\lfloor k/2 \rfloor$ from larger bins, for 10--12 options total.

\textbf{L2: Camera Motion.} From per-view DA3 outputs $(K_i, R_i, t_i)$, we compute the relative pose between two photos. Translation is decomposed into source-camera right / up / forward axes; rotation into yaw and pitch. Each axis is discretised: $|t_\text{axis}| < 0.2$\,m $\rightarrow$ \emph{negligible}; otherwise sign-classified into the directional label. Rotation thresholds: $|\theta| < 10^{\circ}$ $\rightarrow$ \emph{negligible}. The GT is the (translation, rotation) tuple; distractors are alternative tuples randomly subsampled to 12--15 options.

\textbf{L3a: Single-View Perspective Transformation.} Per scene, pick objects $(A, B, C)$ visible in the same view with pairwise separation $\geq 0.5$\,m. Construct a new frame: origin at $A$, forward along $A \to B$ (projected onto the gravity-horizontal plane). Compute $C$'s position in the new frame, then discretise it into 8 horizontal direction bins and 5 distance bins (0.5--1\,m, 1--2\,m, 2--3\,m, 3--5\,m, and more than 5\,m). The GT is the resulting (direction, distance) bin. Distractors are sampled from the remaining direction--distance grid cells, yielding 8--24 options depending on filter strictness and valid distractor availability.

\textbf{L3b: Cross-View Perspective Transformation.} Same as L3a, but $A$ and $C$ must appear in different views; Stage~2 association supplies their cross-view-consistent positions.

\textbf{L4: Spatial Editing.} Per scene, pick $(A, B, C)$ from the object pool with $A \neq C$ and pairwise separation thresholds. Pick an operation: \emph{translate} (along 4 camera-frame horizontals, distance $d = \max(0.5 \cdot d_{AC}, 0.20\,\text{m})$) or \emph{rotate} (around $B$'s vertical axis by $\{90, 180, 270\}^{\circ}$). Apply the edit to $A$, recompute its position relative to $C$, then bin into (8 direction $\times$ 5 distance). The GT must pass the sweet-zone filter ($\pm 10^{\circ}$ in direction; $\geq 0.15$\,m from any distance-bin boundary). Distractors are sampled from the remaining 39 grid cells, total 15 options.

\textbf{L5: Cross-View Visibility Editing.} For each scene-pair (source, target), pick $A$ visible in the source and a set of candidate operations (translations at $\{0.5, 1.0, 1.5, 2.0\}$\,m along 5 directions; rotations around qualifying reference objects at $\{90, 180, 270\}^{\circ}$). For each operation, apply it to $A$ and compute $A$'s vis\_ratio in the target view via Z-buffer rendering against all other objects. Label \emph{visible} ($\text{vis\_ratio} > 0.50$), \emph{invisible} ($< 0.15$), or undefined. The GT operation flips visibility (appear: invisible $\to$ visible; disappear: visible in both $\to$ invisible). Distractors are operations that do not flip visibility; a ``none of the above'' option may be the correct answer.

\paragraph{Question-generation parameter summary.}
L4 translation distance $d = \max(0.5 \cdot d_{AC}, 0.20\,\text{m})$; L4 rotation angles $\{90^{\circ}, 180^{\circ}, 270^{\circ}\}$ around B's vertical axis. Direction sweet zones: $\pm 10^{\circ}$ around each of 8 compass directions, with $25^{\circ}$ dead zones between adjacent directions discarded. Distance bin margin: ground-truth distances within $0.15\,\text{m}$ of any bin boundary are discarded. Candidate-object size filter: minimum edge $0.05\,\text{m}$, maximum edge $0.80\,\text{m}$. SAM3 confidence threshold: $\geq 0.5$. Scene-clutter filter: scenes with more than 15 visible non-structural objects per view are discarded wholesale.

\paragraph{LLM-based question refinement.}
Raw questions produced by Section~\ref{sec:benchmark} contain object identifiers of the form ``obj\_5 (lamp)''. We use Gemini~3.1~Pro to replace each identifier with a natural-language description grounded in visual appearance and position (e.g., ``the red lamp on the left side of the second image''). The LLM receives the three input photos along with object bounding-box coordinates and a nine-cell grid position label. The refinement prompt strictly forbids rewriting, abbreviation, or reorganisation of the question text; only identifier substitution is allowed, preventing semantic drift.

\section{Human Audit Protocol and Statistics}
\label{app:audit}

All 1{,}172 candidate questions were reviewed by multiple annotators against a written protocol. Each question underwent a two-round audit. In the first round, an annotator performed both a stem audit (verifying that the question stem is unambiguous) and an answer-key audit (verifying that the LLM-refined answer is correct), producing one of three outcomes: \textbf{direct pass} (stem and answer key both correct, no editing needed); \textbf{answer-fixed} (stem correct, answer key corrected by the annotator, typically a direction or distance error caused by pipeline-level geometric noise in the spatial-reasoning answer); or \textbf{rejected} (stem itself problematic, most commonly referent ambiguity when multiple visually similar objects co-appear in the same view, or a mis-labelled target object). In the second round, a different annotator independently re-checked the first-round outcome; disagreements were returned for re-review and the cycle iterated until both rounds agreed. Table~\ref{tab:audit-per-task} reports the per-task breakdown; the aggregate ($758$ direct + $245$ answer-fixed + $169$ rejected, also reported in Section~\ref{sec:pipeline-validity}) yields a direct-pass rate of $64.7\%$ and a final acceptance rate of $85.6\%$. Direct-pass rates range from $54.3\%$ (L1) to $72.0\%$ (L3b), with the two counterfactual editing tasks (L4, L5) in the middle at $67.5\%$ and $69.5\%$. Among candidates with valid stems, the LLM-refined answer is already correct in $75.6\%$ of cases (per-task range $66.7\%$--$82.3\%$).

\begin{table*}[t]
  \centering
  \footnotesize
  \setlength{\tabcolsep}{5pt}
  \caption{Per-task pipeline-audit outcomes. \textbf{Cand.}: total candidates from the pipeline. \textbf{Direct}: stem and answer key both verified correct (no editing). \textbf{Fixed}: stem correct, answer key corrected by annotators. \textbf{Rejected}: stem ambiguous or mis-labelled. \textbf{Direct \%} = Direct/Cand. \textbf{Final \%} = (Direct + Fixed)/Cand., the fraction retained in the benchmark.}
  \label{tab:audit-per-task}
  \begin{tabular}{lrrrrrr}
    \toprule
    Task & Cand. & Direct & Fixed & Rejected & Direct \% & Final \% \\
    \midrule
    L1 (Cross-View Obj Rel) & 173 & 94 & 41 & 38 & 54.3 & 78.0 \\
    L2 (Camera Motion) & 200 & 120 & 60 & 20 & 60.0 & 90.0 \\
    L3a (Single-View PT) & 199 & 126 & 41 & 32 & 63.3 & 83.9 \\
    L3b (Cross-View PT) & 200 & 144 & 31 & 25 & 72.0 & 87.5 \\
    L4 (Spatial Editing) & 200 & 135 & 42 & 23 & 67.5 & 88.5 \\
    L5 (Cross-View Vis Edit) & 200 & 139 & 30 & 31 & 69.5 & 84.5 \\
    \midrule
    Total & 1{,}172 & 758 & 245 & 169 & 64.7 & 85.6 \\
    \bottomrule
  \end{tabular}
\end{table*}

\paragraph{Annotators.}
The audit reviews and the human-baseline responses were produced by annotators retained through a professional annotation service. The annotators are familiar with multiple-choice spatial-reasoning task formats. Each candidate question underwent at least two audit rounds (an initial review and a second-round confirmation, with disagreements returned for re-review until both rounds agree), and each human-baseline question was answered independently by three annotators whose responses are aggregated by majority vote. Both tasks followed formal written instructions; the verbatim instructions for the audit decision criteria and for the human-baseline task will be released alongside the benchmark questions. Compensation was provided through the service per its standard rates for visual-reasoning annotation tasks; the authors do not have direct visibility into individual wage rates or the exact number of unique annotators retained by the service. The annotation deliverables may be released together with the MindEdit-Bench benchmark per the service agreement.

\section{Evaluation Protocol}
\label{app:eval-protocol}

This appendix describes how we evaluate both API-served and local VLMs on MindEdit-Bench. All main results use the same 1{,}003 human-verified questions as Section~\ref{sec:experiments}. Each example contains the image paths, the natural-language question, a list of labelled multiple-choice options, and the verified answer letter.

\paragraph{Model interfaces.}
API models are queried through an OpenAI-compatible chat-completion interface. Local models are evaluated with vLLM offline inference using each model's chat template and local image-file inputs. The evaluation runner uses the same message construction and answer extraction logic for both tiers; the main difference is that API requests encode images as data URLs, whereas local vLLM runs pass file URLs to the offline engine.

\paragraph{Decoding and repetition.}
All models are evaluated with greedy decoding (\texttt{temperature=0}) and a maximum generation budget of 2{,}048 tokens. For API models, requests use \texttt{image\_detail=auto}, concurrency 5, a 300-second timeout, and up to three retries with exponential backoff. For local models, vLLM uses the model-specific tensor-parallel and context-length settings, with the same \texttt{temperature=0} and 2{,}048-token generation budget. Every API and local model is run three times on the full benchmark. We report the arithmetic mean across the three runs as the model score and compute the sample standard deviation across runs for run-to-run stability analysis.

\paragraph{Prompt format.}
The user message contains the input images followed by one text block consisting of the question stem, a blank line, and the full option list. This corresponds to \texttt{content\_order=images\_first} and \texttt{no\_images=false} for the main evaluation. The system prompts are listed below. Only \texttt{SYSTEM\_PROMPT} is used for the main benchmark; \texttt{SYSTEM\_PROMPT\_COT} supports diagnostic or supplemental experiments.

\noindent\textbf{System prompt.}\par
\noindent\texttt{SYSTEM\_PROMPT} is the standard multimodal answer-only evaluation prompt.
\begin{lstlisting}[style=prompt,caption={Standard multimodal answer-only system prompt.}]
You are a visual question answering assistant. Carefully inspect the images and answer the multiple-choice question. The option letters may range from A to X. Output only one uppercase option letter, such as A. Do not output reasoning, punctuation, option text, or any other words.
\end{lstlisting}

\noindent\textbf{Multimodal chain-of-thought prompt.}\par
\noindent\texttt{SYSTEM\_PROMPT\_COT} is the multimodal chain-of-thought diagnostic prompt.
\begin{lstlisting}[style=prompt,caption={Multimodal chain-of-thought system prompt.}]
You are a visual question answering assistant. Carefully observe the image and answer the multiple-choice question. First, explain your reasoning step by step, then output your final answer on a separate line in the format: Answer: X (where X is the option letter).
\end{lstlisting}

\paragraph{Scoring and aggregation.}
For each model response, the evaluator extracts a final option letter in the range A--X. Exact matches to the verified answer are counted as correct; responses with no extractable option letter are counted as incorrect. Each completed run writes per-example predictions and raw responses to \texttt{predictions.jsonl}, run-level metrics to \texttt{metrics.json}, and one row in the experiment summary CSV. We aggregate rows by model and compute per-task and overall mean accuracy plus standard deviation across the three runs.

\paragraph{Compute.}
Local-model inference is performed on a single workstation with 2$\times$ NVIDIA A100 GPUs; the full evaluation of the nine open-source models (three runs each on the 1{,}003-question benchmark) takes approximately 24 GPU-hours in total. API-model evaluation cost is approximately USD 400 in aggregate across the six providers.

\section{Per-Model Per-Task Results}
\label{app:full-results}

Table~\ref{tab:main-full} reports per-model, per-task accuracy across all 15 evaluated VLMs.

\begin{table*}[t]
\centering
\footnotesize
\setlength{\tabcolsep}{4pt}
\caption{Per-model, per-task accuracy (\%). Rows sorted by overall mean accuracy. L1 = Cross-View Object Relations; L2 = Camera Motion; L3a = Single-View PT; L3b = Cross-View PT; L4 = Spatial Editing; L5 = Cross-View Visibility Editing. Overall is the pooled accuracy over all 1,003 benchmark examples, using the final per-task example counts in Table 4. It is not the unweighted macro-average of the six task accuracies. The Mean row is the per-task arithmetic average across the 15 models above; this matches the per-task values reported in Section~\ref{sec:main}. Numbers are means across three runs per model; open-source models with non-stochastic decoding produce nearly identical runs.}
\label{tab:main-full}
\begin{tabular}{llrrrrrrr}
\toprule
Model & Type & L1 & L2 & L3a & L3b & L4 & L5 & Overall \\
\midrule
Gemini-3.1-Pro-Preview & API & 58.0 & 28.3 & 31.3 & 18.3 & 36.9 & 40.2 & \textbf{34.6} \\
Qwen3.6-Plus & API & 33.7 & 37.8 & 22.2 & 20.0 & 35.6 & 37.6 & 31.1 \\
Doubao-Seed-2.0-Pro & API & 29.4 & 40.2 & 11.6 & 5.0 & 27.9 & 38.5 & 25.4 \\
Claude-Opus-4-6 & API & 39.8 & 20.9 & 12.8 & 10.5 & 28.4 & 24.7 & 22.2 \\
Qwen3-VL-32B & Local & 41.5 & 21.1 & 10.2 & 7.4 & 15.3 & 23.7 & 19.0 \\
GPT-5.4 & API & 37.0 & 20.4 & 9.8 & 8.2 & 15.3 & 17.2 & 17.3 \\
Qwen3.5-35B-A3B & Local & 39.3 & 15.3 & 9.5 & 5.5 & 20.0 & 16.4 & 16.9 \\
SenseNova-SI-1.1-Qwen3-VL-8B & Local & 31.1 & 15.0 & 13.2 & 9.7 & 18.6 & 12.4 & 16.2 \\
Kimi-K2.5 & API & 23.7 & 25.0 & 3.8 & 2.9 & 19.4 & 21.5 & 15.9 \\
VST-7B-SFT & Local & 20.0 & 19.4 & 15.0 & 12.0 & 15.3 & 7.7 & 14.8 \\
InternVL3-8B & Local & 27.4 & 7.8 & 6.6 & 5.1 & 18.6 & 12.4 & 12.5 \\
RoboBrain2.5-8B & Local & 22.2 & 13.9 & 6.6 & 2.3 & 17.5 & 11.8 & 12.1 \\
SenseNova-SI-1.1-InternVL3-8B & Local & 6.8 & 14.4 & 13.2 & 11.4 & 13.1 & 9.5 & 11.6 \\
Qwen3-VL-8B & Local & 25.9 & 8.9 & 7.2 & 2.3 & 15.8 & 10.1 & 11.2 \\
GLM-4.6V-Flash & Local & 22.5 & 11.1 & 4.0 & 3.5 & 3.5 & 12.4 & 9.0 \\
\midrule
Mean (all models) & --- & 30.6 & 20.0 & 11.8 & 8.3 & 20.1 & 19.7 & 18.0 \\
Random baseline & --- & $\sim$9 & $\sim$7 & $\sim$4 & $\sim$6 & 6.7 & $\sim$10 & --- \\
\bottomrule
\end{tabular}
\end{table*}

\section{Chain-of-Thought Prompting Analysis}
\label{app:cot}

We evaluate the same standard CoT setting by appending a generic step-by-step reasoning instruction to the baseline prompt for four representative models---Gemini-3.1-Pro-Preview, GPT-5.4, Qwen3-VL-32B, and Doubao-Seed-2.0-Pro---over three runs on all 1{,}003 benchmark questions, using the same aggregation methodology as Section~\ref{sec:experiments}. Full per-model per-task results are reported in Table~\ref{tab:cot-comparison}.

Standard CoT yields measurable but task-structure-dependent gains. Across Gemini-3.1-Pro-Preview, Doubao-Seed-2.0-Pro, and Qwen3-VL-32B, CoT improves Group~II perspective transformation tasks (L3a, L3b) by an average of $+3.3$\,pp and Group~III counterfactual editing tasks (L4, L5) by $+3.7$\,pp, while Group~I perception tasks (L1, L2) gain only $+1.8$\,pp. The pattern aligns with task structure: L3 decomposes into ``mentally adopt the new viewpoint, then judge the spatial relation''; L4 into ``apply the virtual transformation, then re-evaluate the direction-distance''; L5 into ``apply the transformation, project to the target view, then check visibility''. CoT prompts that elicit explicit step-by-step reasoning map naturally onto these sequential operations, whereas Group~I tasks involve more holistic visual comparison with less obvious decomposition.

Gains from standard CoT are heterogeneous across models. GPT-5.4 shows the largest CoT improvement ($+12.8$\,pp overall); the other three models show modest gains of $+1.2$\,pp (Gemini-3.1) to $+6.0$\,pp (Qwen3-VL-32B). The small model sample precludes stronger model-level conclusions, and we leave further investigation of model-specific CoT responsiveness to future work.

Despite these gains, standard CoT does not close the human--VLM gap. While individual models benefit from CoT, the benchmark-level SOTA rises only from $34.6\%$ (Gemini baseline) to $35.8\%$ (Gemini + CoT), a gain of just $+1.2$\,pp. The pooled human--VLM gap remains at $52$\,pp; on L3b cross-view perspective transformation, the best CoT-augmented configuration reaches $30.1\%$ versus human $85.7\%$. These results show that generic step-by-step prompting alone is insufficient to close the gap. Although the persistent errors are consistent with limitations in maintaining and manipulating an internal 3D scene representation, this experiment does not isolate that cause or rule out improvements from stronger elicitation strategies or explicit spatial tools.

\begin{table*}[!htbp]
  \centering
  \footnotesize
  \setlength{\tabcolsep}{5pt}
  \caption{Per-model accuracy (\%) under baseline (no CoT) versus standard chain-of-thought (+ CoT) prompting, by task. $\Delta$ rows show CoT minus baseline in percentage points.}
  \label{tab:cot-comparison}
  \begin{tabular}{llrrrrrrr}
    \toprule
    Model & Setting & L1 & L2 & L3a & L3b & L4 & L5 & Overall \\
    \midrule
    \multirow{3}{*}{Gemini-3.1-Pro-Preview}
      & Baseline & 58.0 & 28.3 & 31.3 & 18.3 & 36.9 & 40.2 & 34.6 \\
      & + CoT    & 56.0 & 28.0 & 32.5 & 30.1 & 36.7 & 36.1 & 35.8 \\
      & $\Delta$ & $-2.0$ & $-0.3$ & $+1.2$ & $+11.8$ & $-0.2$ & $-4.1$ & $+1.2$ \\
    \midrule
    \multirow{3}{*}{Doubao-Seed-2.0-Pro}
      & Baseline & 29.4 & 40.2 & 11.6 &  5.0 & 27.9 & 38.5 & 25.4 \\
      & + CoT    & 28.4 & 40.4 & 15.0 &  7.4 & 31.3 & 41.6 & 27.4 \\
      & $\Delta$ & $-1.0$ & $+0.2$ & $+3.4$ & $+2.4$ & $+3.4$ & $+3.1$ & $+2.0$ \\
    \midrule
    \multirow{3}{*}{Qwen3-VL-32B}
      & Baseline & 41.5 & 21.1 & 10.2 &  7.4 & 15.3 & 23.7 & 19.0 \\
      & + CoT    & 47.2 & 29.4 & 11.8 &  7.0 & 24.9 & 34.3 & 25.0 \\
      & $\Delta$ & $+5.7$ & $+8.3$ & $+1.6$ & $-0.4$ & $+9.6$ & $+10.6$ & $+6.0$ \\
    \midrule
    \multirow{3}{*}{GPT-5.4}
      & Baseline & 37.0 & 20.4 &  9.8 &  8.2 & 15.3 & 17.2 & 17.3 \\
      & + CoT    & 46.9 & 33.9 & 20.2 & 14.9 & 27.7 & 40.8 & 30.1 \\
      & $\Delta$ & $+9.9$ & $+13.5$ & $+10.4$ & $+6.7$ & $+12.4$ & $+23.6$ & $+12.8$ \\
    \bottomrule
  \end{tabular}
\end{table*}

\section{Answer Permutation: Per-Task Breakdown}
\label{app:permutation-per-task}
Table~\ref{tab:permutation-per-task} reports the per-task accuracy for each probed model under the original main run and the three option-letter shuffles (mean across the three shuffles), along with the absolute difference $|\Delta|$ between them. Per-task $|\Delta|$ reaches up to 6.9\,pp (Qwen3-VL-32B on L2), 6.4\,pp (SenseNova-SI-1.1-Qwen3-VL-8B on L1), and 5.0\,pp (Gemini-3.1-Pro-Preview on L3b), while the pooled overall accuracy stays within 1.4\,pp on all three models, indicating that per-task fluctuations average out into a position-independent overall signal.

\begin{figure*}[!htbp]
  \centering
  \includegraphics[width=\linewidth]{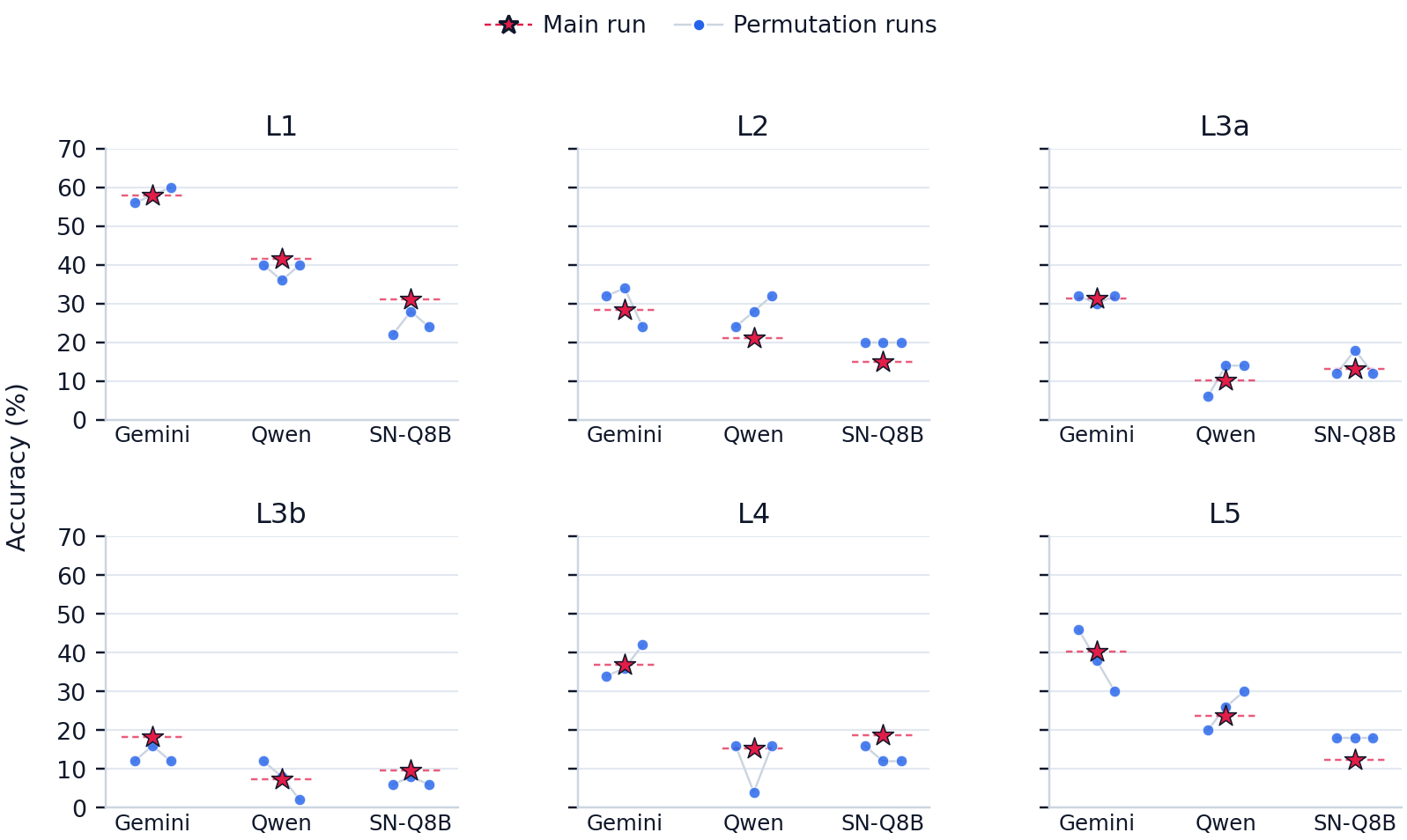}
  \caption{\textbf{Answer-permutation control.} Each panel shows task-level accuracy for three representative VLMs under the main run (rose stars and dashed guides) and three option-letter shuffles (blue points connected by range lines). Reordering option letters changes overall accuracy by at most 1.4\,pp, while the largest task-level fluctuation is 6.9\,pp.}
  \label{fig:permutation}
\end{figure*}

\begin{table*}[!htbp]
  \centering
  \footnotesize
  \setlength{\tabcolsep}{5pt}
  \caption{Per-task answer-permutation breakdown for three probed models. ``Main'' = original main-run accuracy (\%); ``Perm'' = mean accuracy across three option-letter shuffles (\%); ``$|\Delta|$'' = absolute difference between Main and Perm (pp).}
  \label{tab:permutation-per-task}
  \begin{tabular}{llrrrrrrr}
    \toprule
    Model & Row & L1 & L2 & L3a & L3b & L4 & L5 & Overall \\
    \midrule
    \multirow{3}{*}{Gemini-3.1-Pro-Preview}
      & Main & 58.0 & 28.3 & 31.3 & 18.3 & 36.9 & 40.2 & 34.6 \\
      & Perm & 58.0 & 30.0 & 31.3 & 13.3 & 37.3 & 38.0 & 34.7 \\
      & $|\Delta|$ & 0.0 & 1.7 & 0.0 & 5.0 & 0.4 & 2.2 & 0.1 \\
    \midrule
    \multirow{3}{*}{Qwen3-VL-32B}
      & Main & 41.5 & 21.1 & 10.2 & 7.4 & 15.3 & 23.7 & 19.0 \\
      & Perm & 38.7 & 28.0 & 11.3 & 7.3 & 12.0 & 25.3 & 20.4 \\
      & $|\Delta|$ & 2.8 & 6.9 & 1.1 & 0.1 & 3.3 & 1.6 & 1.4 \\
    \midrule
    \multirow{3}{*}{SenseNova-SI-1.1-Qwen3-VL-8B}
      & Main & 31.1 & 15.0 & 13.2 & 9.7 & 18.6 & 12.4 & 16.2 \\
      & Perm & 24.7 & 20.0 & 14.0 & 6.7 & 13.3 & 18.0 & 16.1 \\
      & $|\Delta|$ & 6.4 & 5.0 & 0.8 & 3.0 & 5.3 & 5.6 & 0.1 \\
    \bottomrule
  \end{tabular}
\end{table*}

\section{L5 Meta-Option Selection Rates per Model}
\label{app:l5-fallback}
Table~\ref{tab:l5-fallback} reports each model's selection rate for the L5 meta-option ``none of the above'' alongside its pooled overall accuracy over all benchmark examples. The L5 GT frequency for this option is $14.2\%$; the three weakest open-source models exceed it by a large margin (Qwen3-VL-8B, InternVL3-8B, SenseNova-SI-1.1-Qwen3-VL-8B), while the strongest API models track it closely. RoboBrain2.5-8B selects this option $0\%$ of the time, reflecting an opposite failure mode (refusal to use the fallback at all).

\begin{table}[!htbp]
  \centering
  \footnotesize
  \setlength{\tabcolsep}{6pt}
  \caption{L5 ``none of the above'' meta-option selection rate per model, alongside the model's overall accuracy across the six task types. Models are sorted by meta-option rate (descending). The GT frequency for this option is $14.2\%$. SenseNova-SI-Q3VL-8B and SenseNova-SI-IV3-8B abbreviate the SenseNova-SI-1.1 variants on Qwen3-VL-8B and InternVL3-8B respectively.}
  \label{tab:l5-fallback}
  \begin{tabular}{lrr}
    \toprule
    Model & Meta-opt \% & Overall acc \% \\
    \midrule
    SenseNova-SI-Q3VL-8B          & 95.1 & 16.2 \\
    InternVL3-8B                  & 88.3 & 12.5 \\
    Qwen3-VL-8B                   & 73.0 & 11.2 \\
    Kimi-K2.5                     & 68.9 & 15.9 \\
    Qwen3-VL-32B                  & 44.8 & 19.0 \\
    SenseNova-SI-IV3-8B           & 30.1 & 11.6 \\
    Claude-Opus-4-6               & 18.6 & 22.2 \\
    Doubao-Seed-2.0-Pro           & 14.7 & 25.4 \\
    Gemini-3.1-Pro-Preview        & 14.5 & 34.6 \\
    VST-7B-SFT                    & 12.9 & 14.8 \\
    GPT-5.4                       & 11.3 & 17.3 \\
    GLM-4.6V-Flash                & 6.4  & 9.0 \\
    Qwen3.5-35B-A3B               & 6.3  & 16.9 \\
    Qwen3.6-Plus                  & 1.8  & 31.1 \\
    RoboBrain2.5-8B               & 0.0  & 12.1 \\
    \bottomrule
  \end{tabular}
\end{table}

\section{Ethical Considerations}
\label{app:ethics}

\paragraph{Data collection and privacy.}
The 120 indoor scenes that constitute MindEdit-Bench were privately captured by the authors using consumer smartphones, in spaces that the authors own or for which explicit photography consent was obtained from the space owner. We did not scrape any images from the web or repurpose third-party datasets. Each released image was manually screened during curation to exclude any frame containing human faces, identifying documents, screens displaying personal information, or other personally identifying material. No offensive content is present.

\paragraph{Intended use and misuse risks.}
MindEdit-Bench is intended for academic research on VLM spatial reasoning; broader use is permitted under the license in Section~\ref{sec:conclusion}. The primary misuse risk is over-interpreting scores as evidence of deployment readiness for safety-relevant applications such as assistive navigation or household robotics; the benchmark itself poses no direct dual-use risk.

\paragraph{Use of AI assistants.}
Large language models enter this work in three roles. First, Gemini 3.1 Pro acts as a pipeline component, producing a unified object vocabulary across the three input photos (Section~\ref{sec:pipeline}, Appendix~\ref{app:pipeline}) and refining raw object identifiers into natural-language descriptions under a constrained substitution-only prompt (Appendix~\ref{app:question-gen}). Second, fifteen VLMs are the systems under evaluation (Section~\ref{sec:setup}). Third, during manuscript preparation, large language models (ChatGPT, Claude) were used solely for English-language polishing; they did not contribute to the core methodology, experimental design, data collection, or analysis.

\end{document}